\documentclass[letterpaper, 10 pt, journal, twoside]{IEEEtran}

\usepackage{graphicx}

\usepackage{etoolbox}
\makeatletter
\patchcmd{\@makecaption}
  {\scshape}
  {}
  {}
  {}
\makeatletter
\patchcmd{\@makecaption}
  {\\}
  {.\ }
  {}
  {}
\makeatother
\def\tablename{Table}

\usepackage{times}
\usepackage{booktabs}

\usepackage[noend]{algpseudocode}
\usepackage{algorithm}
\usepackage{amsmath}
\usepackage{amsfonts}
\usepackage{bm}
\usepackage{comment}

\newcommand{\new}[1]{#1}

\newcommand{\gray}[1]{\textcolor{gray}{#1}}
\newcommand{\xs}{\underline{x}}
\newcommand{\link}[1]{\href{#1}{#1}}

\usepackage{tikz}
\usetikzlibrary{bayesnet}

\usetikzlibrary{arrows.meta}
\tikzset{%
  >={Latex[width=2mm,length=2mm]},
            base/.style = {rectangle, rounded corners, draw=black,
                           minimum width=1cm, minimum height=1cm,
                           text centered, font=\small},
  activityStarts/.style = {base, fill=blue!30},
       startstop/.style = {base, fill=red!30},
    activityRuns/.style = {base, fill=green!30},
         inout/.style = {base, minimum width=6cm, minimum height=0cm,
                           font=\small},
         process/.style = {base, minimum width=1.5cm, 
                           font=\small},
         data/.style = {base,minimum width=1.5cm, 
                           font=\small},
         sec/.style = {base, minimum width=1.5cm, 
                           font=\color{gray}\small, draw=none,
                           fill=none },
         processX/.style = {base, minimum width=1.8cm, 
                           inner sep=1pt,
                           minimum height=.5cm,
                           font=\footnotesize},
         processR/.style = {processX, fill=red!30},
         processG/.style = {processX, fill=green!30},
         number/.style = { minimum width=1cm, minimum height=0cm,
                           text centered, font=\small},
}

\newcommand{\RR}{\mathbb{R}}

\newcommand{\seq}[2]{ \ensuremath{\langle #1 \ldots #2 \rangle }}

\newtheorem{property}{Property}[section]
\newtheorem{proposition}{Proposition}[section]
\newtheorem{definition}{Definition}[section]
\newtheorem{theorem}{Theorem}[section]

\usepackage{multicol}
\usepackage[bookmarks=true]{hyperref}

\graphicspath{{pics/}{/home/quim/Downloads/videxp/low/clean/} {/home/quim/stg/lgp-pddl/03-pddlSolver/}}

\IEEEoverridecommandlockouts                              

\title{\LARGE \bf
 A Conflict-driven Interface between Symbolic Planning \\ and Nonlinear Constraint Solving
}

\author{Joaquim Ortiz-Haro$^{1}$ $\quad$  Erez Karpas$^{2}$ $\quad$ Michael Katz$^{3}$ $\quad$ Marc Toussaint$^{1}$
\thanks{Manuscript received: February, 24, 2022; Revised May, 24, 2022; Accepted June, 26, 2022.}
\thanks{This paper was recommended for publication by Editor Hanna Kurniawati upon evaluation of the Associate Editor and Reviewers' comments.}
\thanks{This research has been supported by the German-Israeli Foundation for Scientific Research (GIF) grant I-1491-407.6/2019. Joaquim Ortiz-Haro thanks the International Max-Planck Research School for Intelligent Systems (IMPRS-IS) for the support.}
\thanks{$^{1}$ TU Berlin, Germany. }
\thanks{$^{2}$  Technion, Israel. } 
\thanks{$^{3}$ IBM Research, USA.  } 
\thanks{Digital Object Identifier (DOI): see top of this page.}
}

\begin{document}

\maketitle


\begin{abstract}

  Robotic planning in real-world scenarios typically requires joint optimization of logic and continuous variables. 
  A core challenge to combine the strengths of logic planners 
  and continuous solvers is 
  the design of an efficient interface that informs the logical search about continuous infeasibilities. 
  In this paper we present a novel iterative algorithm that connects logic planning with nonlinear optimization through a bidirectional interface, achieved by the detection of minimal subsets of nonlinear constraints that are infeasible.
  \new{
  The algorithm continuously builds a database of graphs that represent (in)feasible subsets of continuous variables and constraints,
}
and encodes this knowledge in the logical description. 
  As a foundation for this algorithm, we introduce \textit{Planning with Nonlinear Transition Constraints (PNTC)}, \new{a novel planning formulation that clarifies the exact assumptions our algorithm requires and can be applied to model Task and Motion Planning (TAMP) efficiently.}
Our experimental results show that our framework significantly outperforms alternative optimization-based approaches for TAMP.

%
\noindent Webpage: \link{https://quimortiz.github.io/graphnlp/}

\end{abstract}


\section{Introduction}
\IEEEPARstart{R}{obot} planning involves both discrete and continuous decisions. For example, in Task and Motion Planning (TAMP,
\cite{garrett2021integrated}), 
discrete decisions concern which
type of interactions with which objects are to be performed,
%
typically formalized using a logic planning language such as STRIPS or PDDL. Continuous decisions concern robot \& object poses, motions and potentially forces
\cite{20-toussaint-RAL},  which need to respect geometric, physical and collision constraints, according to the discrete decisions. The focus of this paper is
the case where continuous constraints are formulated as a nonlinear mathematical program (NLP)
over continuous variables \cite{toussaint2015logic}.



Despite recent advances in TAMP solvers, current algorithms struggle in high dimensional configuration spaces (e.g. several robots), large symbolic spaces and constrained environments that require joint optimization.
A promising approach to plan in such challenging settings is to efficiently interface state-of-the-art solvers on both sides, in particular, incorporating information about (in)feasibility from continuous solvers back to the logical level.
Looking into similar challenges in classical planning formulations, such as Satisfiability Modulo Theory (SMT \cite{de2011satisfiability}), a fundamental approach to inform and guide the logical search is to automatically identify and block \emph{minimal conflicts} which guarantee infeasibility of the continuous problem. 

In this paper we present the \textit{Graph-NLP Planner (GNPP)},  a novel method to interface a logic solver with an NLP solver, which iteratively detects \emph{minimal infeasible subgraphs}, i.e., minimal subsets of nonlinear constraints that are infeasible, and encodes back this information into the logic problem description. The algorithm continuously builds a database of (in)feasible subgraphs -- in a sense learning what is feasible or infeasible -- and uses this accumulated information to inform logic search as well as avoid future feasibility checks.

\begin{figure}[t]

  \centering
  \begingroup
\setlength{\tabcolsep}{2pt} 

  \begin{tabular}{cc}
    






    \includegraphics[width=.45\linewidth]{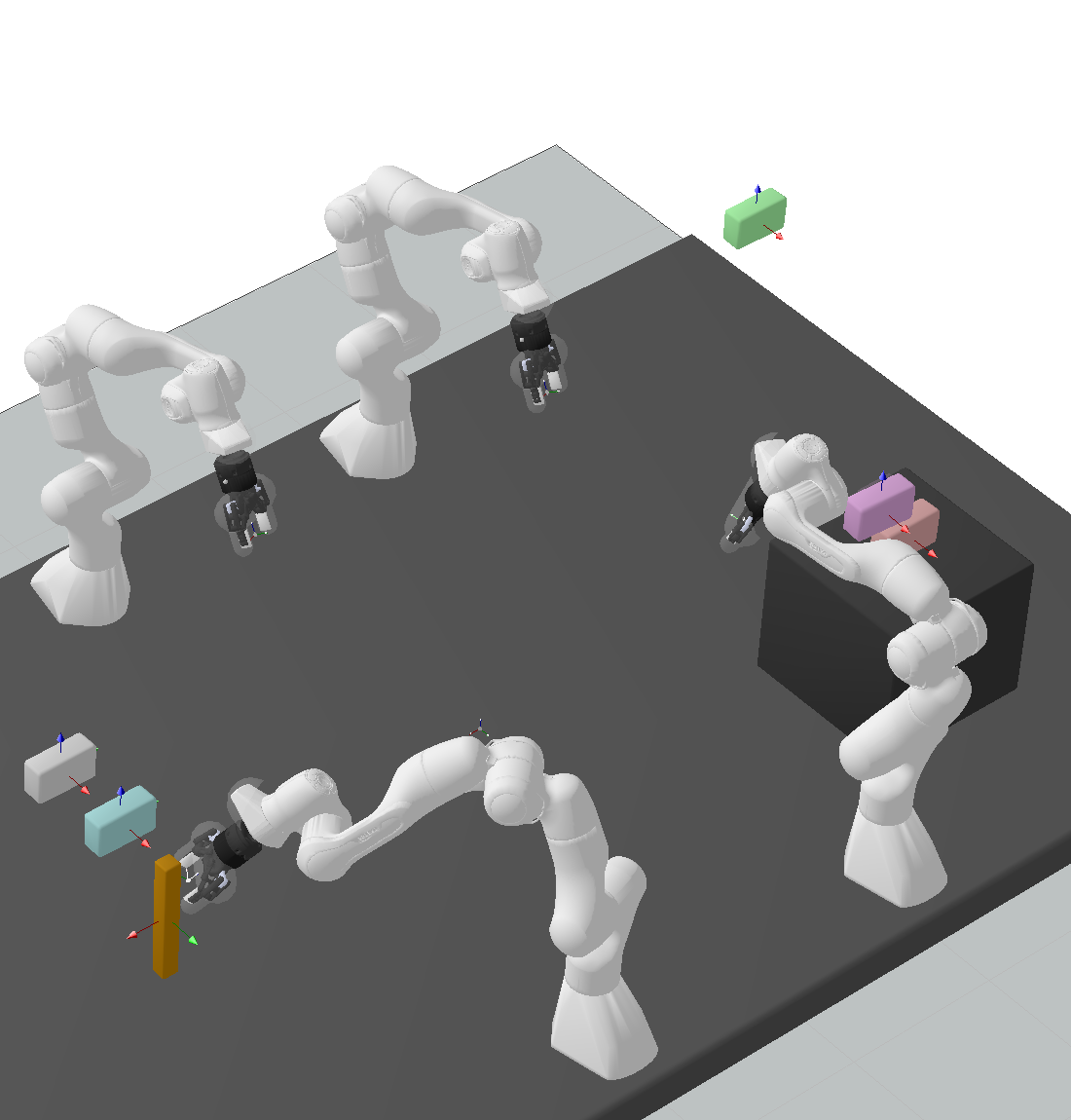}&
    \includegraphics[width=.45\linewidth]{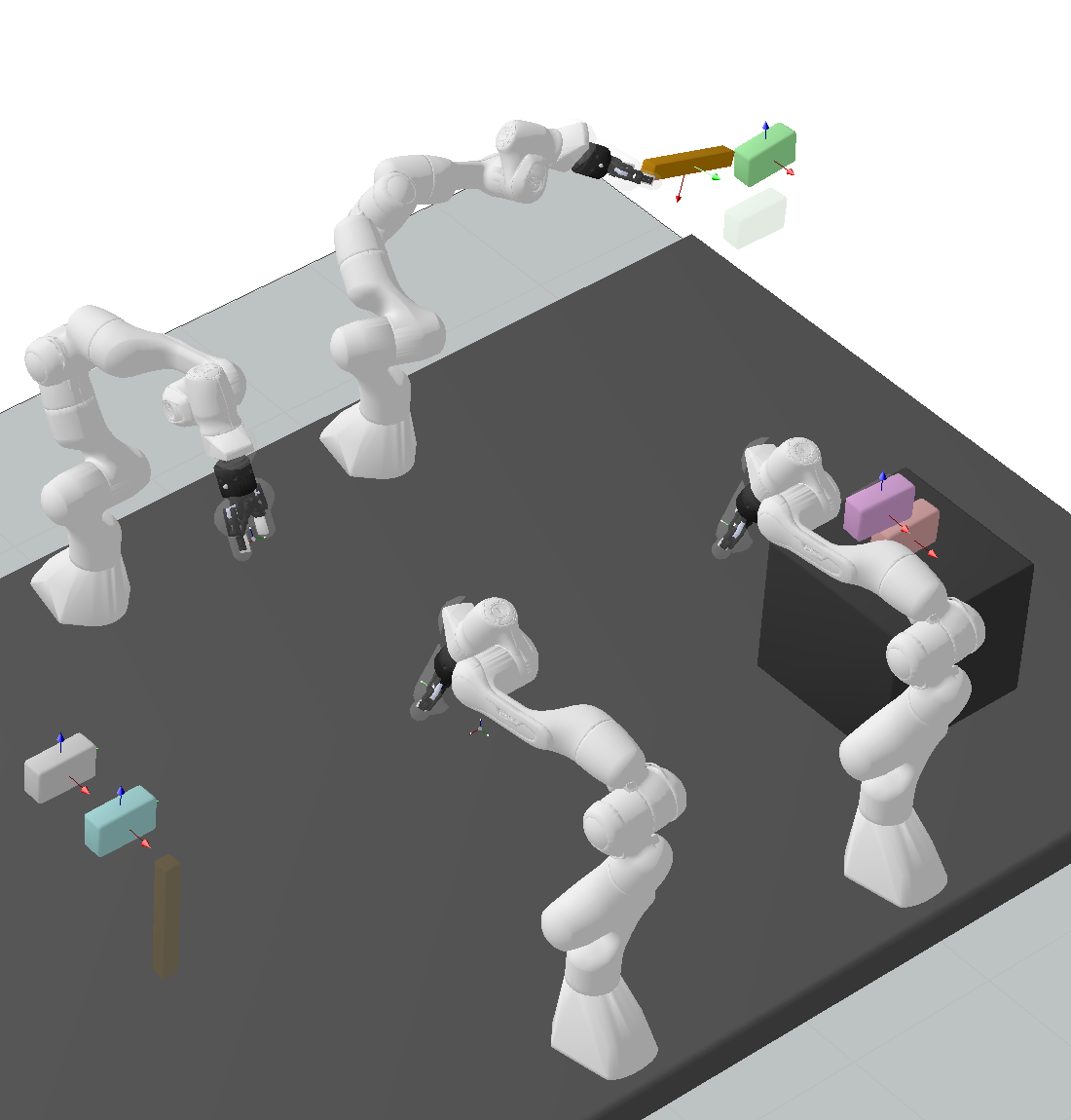} \\

    \includegraphics[width=.45\linewidth]{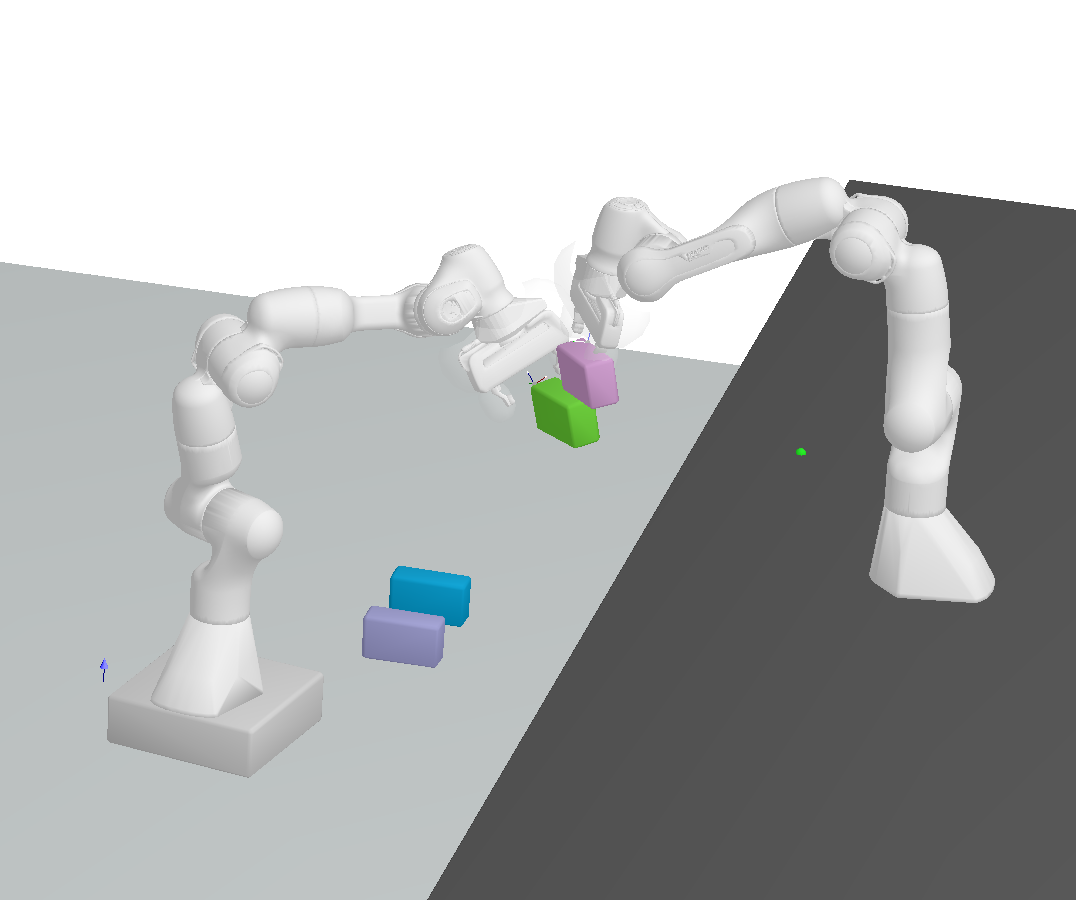}&
    \includegraphics[width=.45\linewidth]{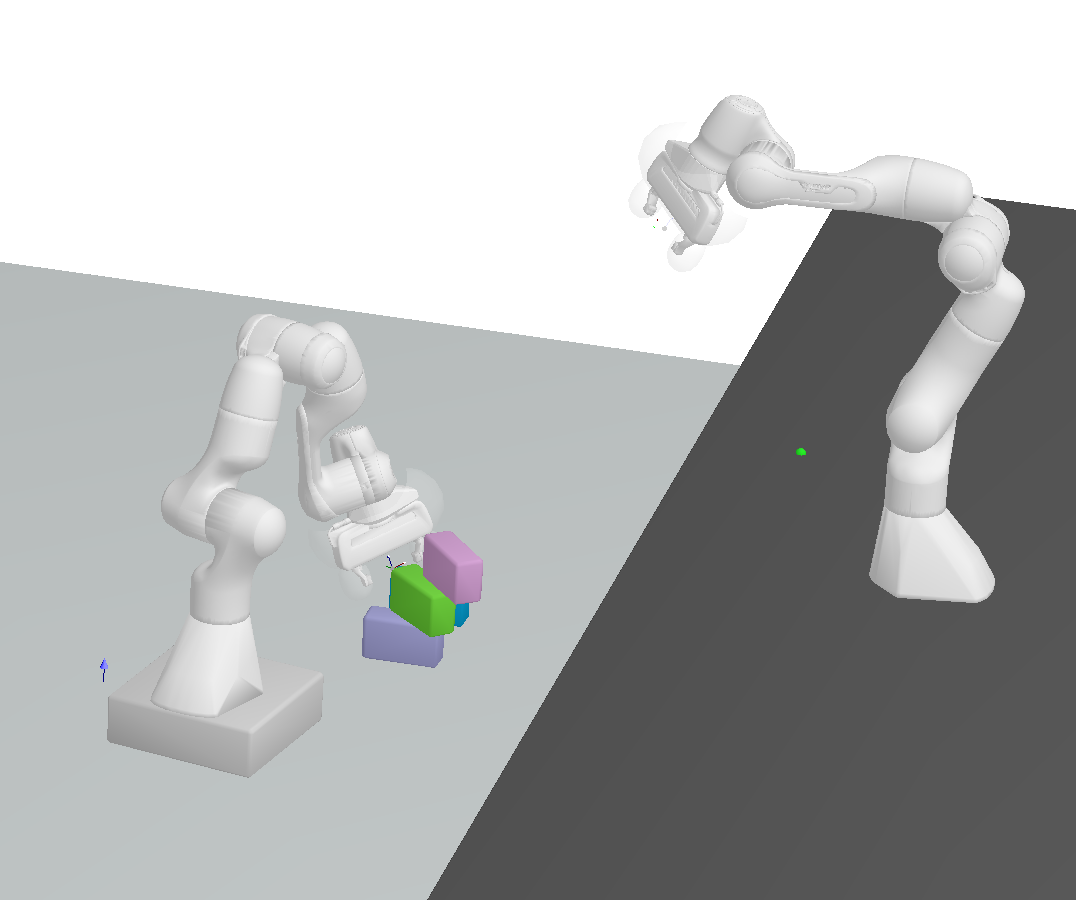} \\

    \includegraphics[width=.45\linewidth]{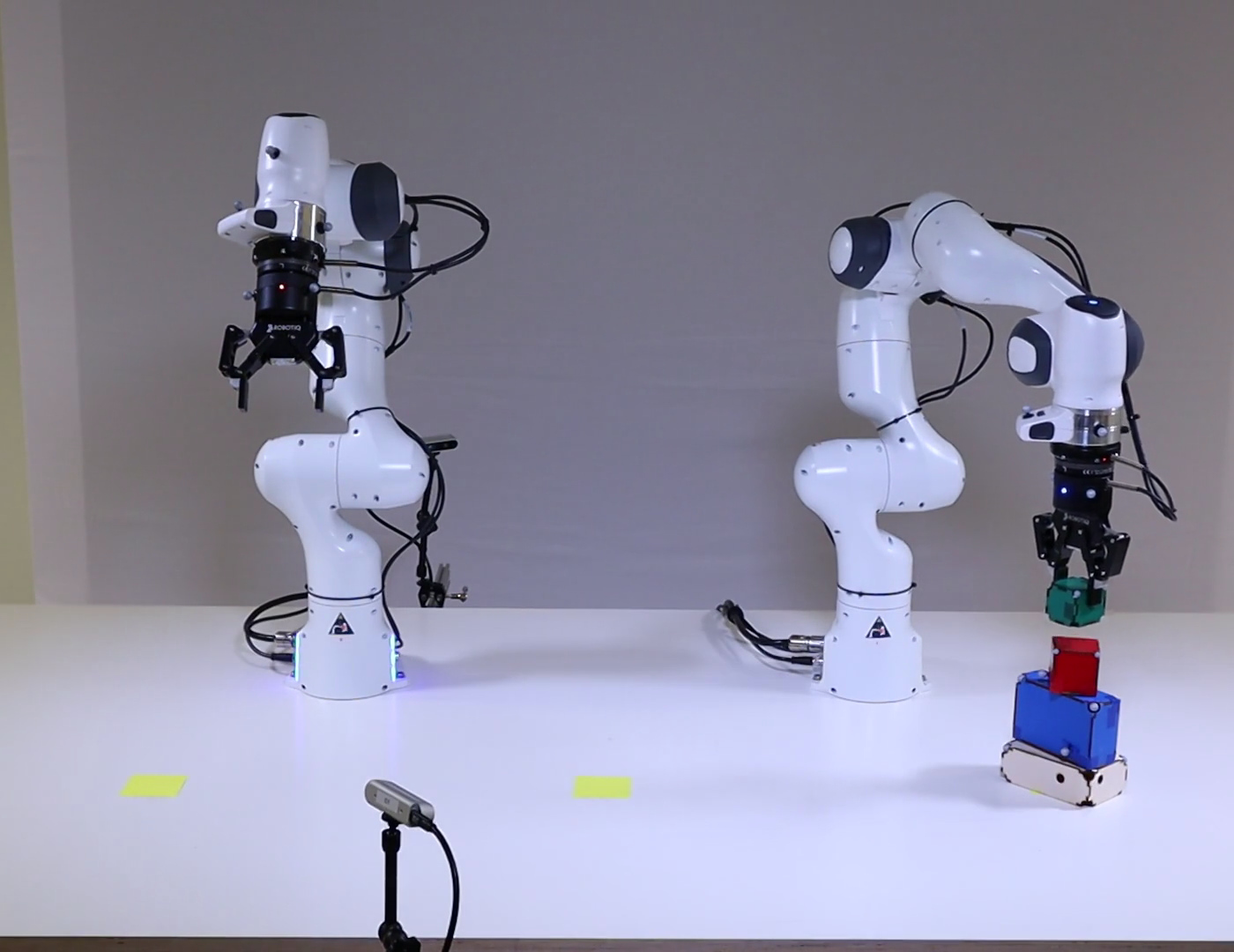} &
    \includegraphics[width=.45\linewidth]{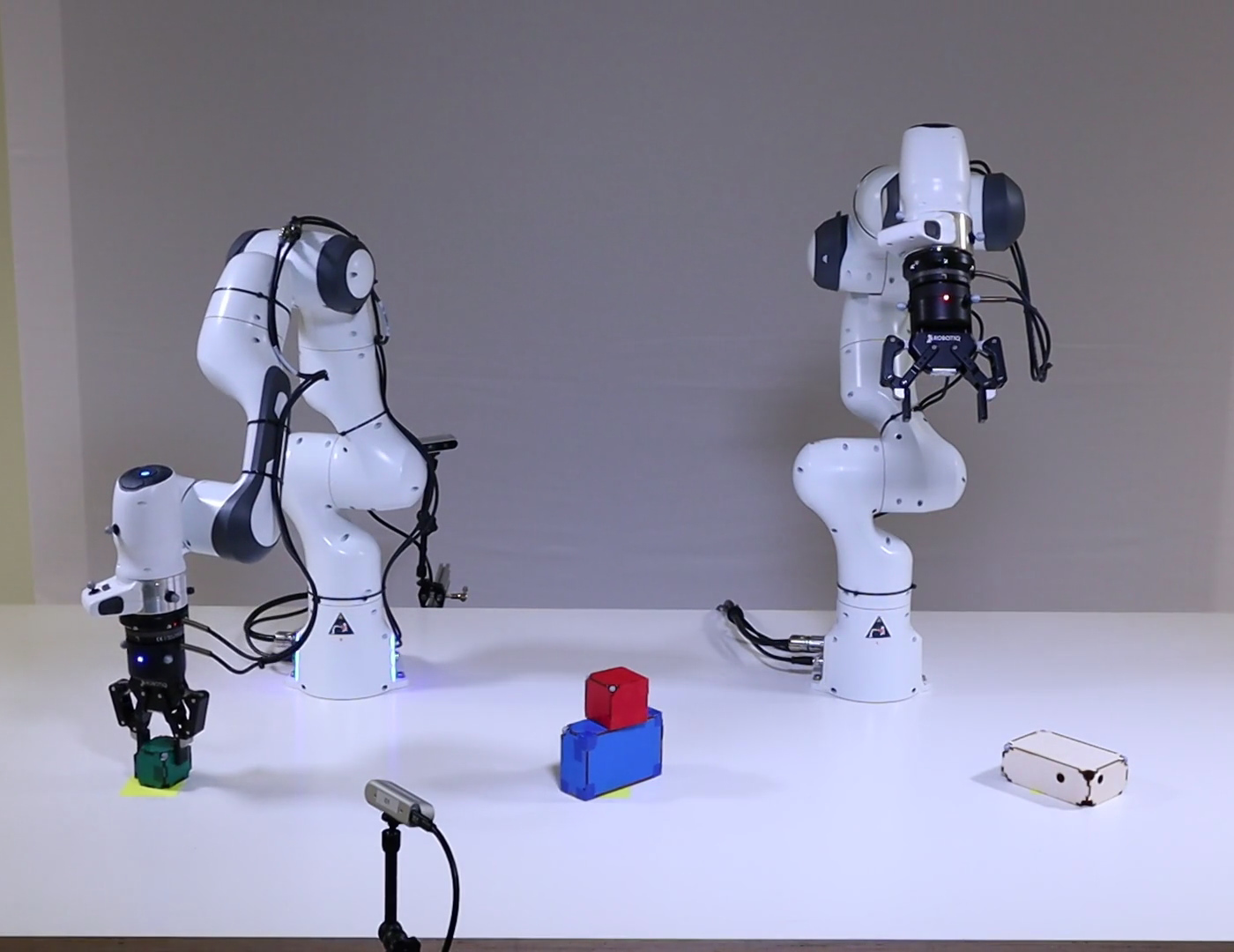} \\

  \end{tabular}
  \endgroup




  \caption{Task and Motion Planning problems solved by our framework. \textit{Top row}: four robot manipulators use a stick as a tool to reach a distant block.
\textit{Mid row}: a heterogeneous team of robots builds a tower. \textit{Bottom row}: two real 7-DOF manipulators solve the Tower of Hanoi puzzle.}
  \label{fig:showcase}
\end{figure}

As a foundation for this algorithm we introduce an abstract \textit{Planning with Nonlinear Transition Constraints (PNTC)} 
formulation, where a symbolic plan implies a factored non-linear program as a sub-problem, and logical predicates of the plan can be related to factors of the NLP. 
This formulation clarifies the concepts and exact assumptions our algorithm builds on, and formally defines 
the explicit bidirectional relation between the logical and the continuous components of the problem, which is
exploited in our solver.

\new
{
PNTC provides a natural and efficient formulation of TAMP problems in robotics that can be viewed as a variant of a Logic-Geometric Program (LGP) \cite{toussaint2015logic}. However, in comparison to LGP, PNTC explicitly defines a factored structure of the implied NLP and a bidirectional mapping between symbols and constraint factors in the resulting NLP. This is exactly the structure we need
  to inform better the symbolic search and is naturally available in TAMP, making our solver directly applicable to solve TAMP and LGP problems.
}
Our method is evaluated on three robotic TAMP scenarios with complex intrinsic logic-geometric dependencies that require long action sequences, generating a solution in a few seconds -- state-of-the-art performance. We deploy the framework in real-world experiments and demonstrate that the solver computes full task and motion plans in real-time.

\section{Related Work}

\subsection{Planning with continuous variables}

\new{
Classical AI planners have been extended to support planning
with numerical constraints on action preconditions (e.g., Metric-FF \cite{koehler1998planning}).
Recent versions of the Planning Domain Definition Language (PDDL)
include temporal planning with numerical variables \cite{fox2006modelling,piotrowski2016heuristic,scala2016interval}. 
}
For instance, 
the COLIN planner \cite{DBLP:journals/jair/ColesCFL12} includes continuous linear change of numerical variables (e.g., fixed velocities), and encodes the temporal and state evolution constraints implied by a sequence of actions as a linear program.
The Scotty planner \cite{fernandez2018scottyactivity} adds support for control variables which govern the rate of change (e.g., controlling the velocity) by replacing the linear program with a Second Order Cone Program. 
Perhaps the most closely related work is \cite{haslum2018extending}, that extends classical planning with general state constraints.  
In contrast, the nonlinear constraints of \textit{PNTC} are defined by a sequence of symbolic states and evaluated on consecutive continuous variables. 
This implies a  nonlinear program for the whole sequence of continuous variables, that must be solved jointly, and can model efficiently long-term relationships between the continuous variables without additional discretization. 




\subsection{Task and Motion Planning} 

Task and Motion Planning in robotics is a prominent example for jointly solving for discrete and continuous variables.  Most TAMP solvers rely on a discretization of the configuration space.
For instance, PDDLStream \cite{garrett2020pddlstream} uses constrained samplers for generating grasp and kinematic solutions inside PDDL-like planning; and \cite{ferrer2017combined} integrates samples of feasible configurations into the planning task through a precompilation.

\new{
Some sampling-based TAMP solvers reason explicitly about geometric conflicts.
For example, a set of predefined predicates such as ``is reachable" is used in  \cite{srivastava2014combined} to combine a black-box task planner with a motion planner. The constraint based approach in \cite{dantam2016incremental} incorporates information about geometric infeasibility by blocking the full task plan or, in some special cases, a pair of a (partial) state and an action. Alternatively, our framework can encode continuous infeasibilities that potentially involve several motion phases. In fact, instead of enumerating possible geometric failure cases, 
we define nonlinear constraints to model the motion and geometry, and let the solver detect which intrinsic subset is jointly infeasible. 
}


Our method provides an optimization-based formulation of TAMP \cite{migimatsu2020object}, \cite{zhao2021sydebo}, \cite{toussaint2018differentiable, ortiz22conflict} which leverages nonlinear optimization to jointly compute a motion that satisfies all geometric and physical constraints. In contrast to previous solvers for LGP \cite{toussaint2018differentiable}, namely Multi-Bound Tree Search \cite{toussaint2017multi},
our solver provides a more efficient logic-geometric interface based on detecting infeasible subsets of constraints instead of infeasible action sequences, as our evaluations show.

\section{Problem Formulation}
\label{sec:problem-formulation}

A \textit{Planning with Nonlinear Transition Constraints (PNTC)} problem is a 7-tuple $\langle \mathcal{V}, \mathcal{A},s_0,g, \Pi, \mathcal{H}, \mathcal{X} \rangle$. 
The logical component $\langle \mathcal{V}, \mathcal{A},s_0,g\rangle$ corresponds to a classical planning problem encoded in SAS+ \cite{backstrom-nebel-compint1995}. 
$\mathcal{V}$ is a finite set of variables and $\mathcal{A}$ is a finite set of action operators.
Each variable $v \in \mathcal{V}$ has a finite domain $dom(v)$. A symbolic state $s$ is an assignment to the variables $v \in \mathcal{V}$. A partial state $p$ is an assignment to a subset of variables. 
$s_0$ is the initial state and $g$ is a partial state that represents the goal.
We denote by $\mathcal{P} = \times_{v \in \mathcal{V}} ( dom(v) \cup \{\bot\})$ the space of partial states, where $\bot$ means that the partial state does not instantiate a given variable. 
Each action operator $a \in \mathcal{A}$ is a pair of partial states called preconditions and effects $\langle \text{pre}(a), \text{eff}(a) \rangle$. An action $a$ is applicable in state $s$ if $pre(a) \subseteq s$ and modifies variables in $ v \in \text{eff}(a)$. $s' = s[a]$ denotes the resulting state after applying action operator $a$ on $s$. 


\new{$\mathcal{X}$ is a finite set of continuous variables $\{x^1 \ldots x^K\}$. Each variable takes value in a continuous space $X^k$ (e.g. $\text{dom}({x^k}) = X^k = \RR^{n_k}$),
  A continuous state $\underline{x}=\{\underline{x}^1 \ldots \underline{x}^K\}$ is a value assignment to $\{x^1 \ldots x^K\}$.
%
$\mathcal{H}$ is a finite set of nonlinear piece-wise differentiable constraint functions that are evaluated on pairs of subsets of continuous variables,
$\mathcal{H} = \{  h_b :  X^{b_{0}} \times X^{b_{1}}   \to \RR^{n_b}    \}$.
The index sets $b_0,b_1 \subseteq \{1\ldots K\}$ indicate on which subsets of variables the function $h_b$ depends on.
These functions define nonlinear constraints $h \le 0$ on two consecutive subsets of continuous variables $(x^{b_0} , x'^{b_1})$.}

The logical and continuous components of a \textit{PNTC} are coupled through the mapping $\Pi: \mathcal{P} \times \mathcal{P} \to \mathcal{H} \cup \emptyset $, that maps consecutive partial states $\langle p,p' \rangle$ to a nonlinear constraint function $h_b$ evaluated on $\langle x^{b_0}, x'^{b_1}\rangle$, i.e. $\Pi: \langle p,p'\rangle \mapsto h_{b}(x^{b_0},x'^{b_1})$, (the empty set $\emptyset$ highlights that some $
\langle p,p' \rangle $ do not generate constraints).
This formulation includes dimension-reducing constraints $h = 0$ (rewritten as $h \le 0$ , $-h \le 0$) and constraints acting on a single state $\Pi(p) \to h_b(x^{b_0})$.












A solution is a sequence of logical and continuous states $\langle (s_0,
\xs_0) \ldots (s_n,\xs_n)  \rangle$ and action operators $\langle a_1 \ldots a_n \rangle$ such that $s_i = s_{i-1}[a_i]$, $g \subseteq s_n$ and
$h_{b}(\xs_i^{b_1},\xs_{i+1}^{b_0}) \le 0, ~ h_b = \Pi(p,p')$, $\forall p \subseteq s_i, p' \subseteq s_{i+1}, \forall i=0\ldots n-1$ (using $\xs_i = \{\xs_i^1 \ldots \xs_i^K \} $).
Given a fixed logical plan \seq{s_0}{s_n} the continuous states can be computed by solving the 
continuous feasibility program, i.e. nonlinear program without costs:
\new{
\begin{align}
  \label{eq:nlp}
& \text{find}     ~ x_i^k \in X^k, ~ \forall i= 0 \ldots n,~ k=1\ldots K \\
& \text{s.t}  \ h_{b}(x_i^{b_0},x_{i+1}^{b_1}) \le 0, ~ h_b = \Pi(p,p') \nonumber \\
   & \quad \quad \forall p \subseteq s_i, p' \subseteq s_{i+1}, \forall i=0\ldots n-1 \nonumber
\end{align}
}
Therefore, a valid logical plan is only a necessary condition 
for the existence of the full logical and continuous solution and, in practice, valid logical plans often fail at the continuous level. 

\section{Graph-NLP: a Bidirectional Logic-continuous Interface}
\label{sec:graphnlp}

 

Given a fixed sequence of logical states $\langle s_0  \ldots  s_n \rangle$, we represent the NLP on the continuous variables $x_i^k$  \eqref{eq:nlp} as a graph-NLP, a structured representation that resembles constraint graphs \cite{dechter1992constraint}, graphical models \cite{koller2009probabilistic} or factor graphs.
 We now state the definition and some properties that will later be exploited by our algorithms.

\begin{definition}
  A graph-NLP $G(\langle s_0 \ldots s_n \rangle) =(V_G,E_G)$ is a bipartite graph that models continuous variables and constraints (vertices) and their dependencies (edges) for the fixed sequence of logic states $\langle s_0 \ldots s_n \rangle$. Formally,  $V_G=X_G \cup H_G $, where $X_G= \{ x_i^k , ~ i \in 0\ldots n,~ k \in 1 \ldots K \}$ and $H_G = \{ h_{b}: h_b = \Pi(p,p') ~ \forall p \subseteq s_i, p' \subseteq s_{i+1}, \forall i=0 \ldots n-1 \}$,
and $E_G = \{ ( x_i^k, h_b ):
~ h_b \in H_G ~
\text{depends on } ~
x_i^k \in X_G   \}$.

\end{definition}

In relevant applications, such as TAMP,  each constraint $h \in H_G$ depends only on small subsets of variables, which results in sparse and factored graph-NLPs (e.g. Fig. \ref{fig:cg_example2}).


A subset of variables and constraints is a subgraph-NLP:
\begin{definition}
  A subgraph-NLP $M \subseteq G$ of a graph-NLP $G=(X_G\cup H_G , E_G)$ is  $M = (X_M \cup H_M, E_M) $ with $X_M \subseteq X_G $,   $H_M  \subseteq \{ h \in H_G ~ : \text{Neigh}( h ) \subseteq X_M \}$.
\end{definition}

A graph-NLP is locally time connected, and the factors are time invariant.

\begin{property}
  \label{property:localtime}
  (Local time connectivity)
  A variable vertex $x_i^k$ is connected to constraints that are evaluated on
  variables with time index $i$, $i-1$ or $i+1$.
\end{property}

\begin{property}
  (Factor time invariance)
A sequence of partial states induces a subgraph $M(\seq {p_0} {p_L}) = (X_M\cup H_M, E_M)$, with $H_M = \{h_b: h_b = \Pi(p,p') ~ \forall p\subseteq p_l , p' \subseteq p_{l+1}, ~ l \in 0\ldots L-1 \}$ and $X_M = \{ x^k_l, ~ l\in 0 \ldots L ,~  k  \in 1 \ldots K: ~ \exists h \in H_M  :$ $ ~ h ~ \text{depends on} ~ x^k_l   \}$. 
\end{property}

\begin{property}
  \label{prop:subgraph}
  If \seq{s_0}{s_n} contains \seq {p_0} {p_L} (i.e. $ \exists  i \in \{0 \ldots n-L\} : p_l \subseteq s_{i+l} ~ \forall l=0\ldots L$), then
$M(\seq{p_0}{p_L}) \subseteq G(\seq{s_0}{s_n})$.
\end{property}

\begin{definition}
  A graph-NLP $G$ is said to be feasible $(\text{Feas}(G)=1)$ if there exists a variable assignment $ \xs_i^k, ~ \forall x_i^k \in X_G $  such that all constraints $h \in H_G$ are satisfied. Otherwise $G$ is infeasible ($\text{Feas}(G)=0$). Note that any subgraph $M \subseteq G$ can be evaluated for feasibility. 

\end{definition}

  An assignment can be computed with nonlinear constrained optimization methods such as interior points or augmented Lagrangian (used in our implementation).
  \new{
  The computational complexity (determined by the factorization of a banded diagonal Hessian matrix) is $O(n K^3)$.}
\begin{property}[Monotone Infeasibility]
  \label{propo:monotone} If a subgraph-NLP $M \subseteq G$ is infeasible, $G$ is infeasible.
  If $G$ is a feasible graph, $M \subseteq G$ is feasible. 
\end{property}










\begin{definition}
An infeasible subgraph-NLP  is minimal if, when removing one or more variables or constraints, the resulting graph is feasible.
\end{definition}
\begin{property}
\label{propo:minimal-connected}
The minimal infeasible subgraph is connected. If the graph-NLP $G$ is not connected, the NLP associated to each connected component $G_i$ can be solved independently and $\text{Feas}(G) = \bigwedge_i  \text{Feas}(G_i)$. 
\end{property}

\subsection{Example domain}
\label{sec:example}

\begin{figure}
  \centering
\begin{tikzpicture}[scale=0.7,every node/.style={transform shape}]

        \node[latent] (a0) {$a_0$} ;

        \node[latent,above=.5  of a0,draw=gray] (taua0) {
            \gray{$\tau^a_0$}} ;

        \node[latent,below=.5 of a0  ] (b0) {$b_0$} ;
        \node[latent,below=.5  of b0] (q0) {$q_0$} ;
        \node[latent,below=.5  of q0] (w0) {$w_0$} ;

        \node[latent,below=.5  of w0,draw=gray] (tauw0) {

          \gray{ $\tau^w_0$}} ;


        \node[latent,right=2 of a0  ] (a1) {$a_1$} ;
        \node[latent,below=.5 of a1  ] (b1) {$b_1$} ;
        \node[latent,below=.5 of b1] (q1) {$q_1$} ;
        \node[latent,below=.5 of q1] (w1) {$w_1$} ;
        \node[latent,above=.5  of a1,draw=gray] (taua1) {\gray{$\tau^a_1$}} ;

        \node[latent,below=.5  of w1,draw=gray] (tauw1) {\gray{$\tau^w_1$}} ;

        \node[latent,right=2 of a1  ] (a2) {$a_2$} ;
        \node[latent,below=.5 of a2  ] (b2) {$b_2$} ;
        \node[latent,below=.5 of b2] (q2) {$q_2$} ;
        \node[latent,below=.5 of q2] (w2) {$w_2$} ;

        \node[latent,above=.5  of a2,draw=gray] (taua2) {\gray{$\tau^a_2$}} ;

        \node[latent,below=.5  of w2,draw=gray] (tauw2) {\gray{$\tau^w_2$}} ;

        \node[latent,right=2 of a2  ] (a3) {$a_3$} ;
        \node[latent,below=.5 of a3  ] (b3) {$b_3$} ;
        \node[latent,below=.5 of b3] (q3) {$q_3$} ;
        \node[latent,below=.5 of q3] (w3) {$w_3$} ;


      \factor[left=1 of w0, yshift=0.5cm] {trajp0} { Ref } {w0} {};
      \factor[left=1 of q0, yshift=0.5cm] {trajp0} { Ref } {q0} {};
      \factor[left=1 of a0, yshift=0.5cm] {trajp0} { Ref } {a0} {};
      \factor[above=.3 of a1,xshift=-.5cm] {trajp0} { left:Ref } {a1} {};
      \factor[above=.3 of a2, xshift=-.5cm] {trajp0} { left:Ref } {a2} {};

      \factor[left=1 of b0, yshift=0.5cm] {trajp0} { Ref } {b0} {};
      \factor[left=1 of b1, yshift=0.5cm] {trajp0} { below:Grasp } {b1} {};
      \factor[left=1 of b2, yshift=0.5cm] {trajp0} { Grasp } {b2} {};
      \factor[right=1 of b3, yshift=0.5cm] {trajp0} { Pos } {b3} {};
      \factor[above=.3 of a3,xshift=-.5cm] {trajp0} {left:Ref} {a3} {};




      \factor[left=1 of b1, yshift=-.5cm] {trajp0} {below:Kin} {b0, q1,b1} {};

      \factor[right=1.2 of b1, yshift=-.3cm] {trajp0} {Kin} {b1, q2,b2,w2} {};

      \factor[left=.7 of b3, yshift=-.3cm] {trajp0} {Kin} {w3,b3,a3,b2} {};

      \factor[left=1 of a3] {trajp0} {Equal} {a2,a3} {};

      \factor[left=1 of a1] {trajp0} {Equal} { a0, a1} {};

      \factor[left=1 of a2] {trajp0} {Equal} { a1, a2} {};

      \factor[right=.3 of a0, yshift=-0.5cm,color=brown] {} {} {a0,b0} {};
      \factor[right=.3 of b0, yshift=-0.5cm,color=brown] {} {} {b0,q0} {};
      \factor[right=.3 of q0, yshift=-0.1cm,color=brown] {} {} {a0,q0} {};

      \factor[right=.3 of q0, yshift=-0.5cm,color=brown] {} {} {q0,w0} {};
      \factor[left=.3 of w0, yshift=-0.5cm,color=brown] {} {} {a0,w0} {};
      \factor[left=.3 of w0, yshift=+1cm,color=brown] {} {} {w0,b0} {};

      \factor[right=.3 of a1, yshift=-0.5cm,color=brown] {} {} {a1,b1,q1} {};
      \factor[right=.3 of b1, yshift=-0.5cm,color=brown] {} {} {b1,q1} {};
      \factor[right=.3 of q1, yshift=-0.1cm,color=brown] {} {} {a1,q1} {};

      \factor[right=.3 of q1, yshift=-0.5cm,color=brown] {} {} {q1,w1} {};
      \factor[left=.3 of w1, yshift=-0.5cm,color=brown] {} {} {a1,w1} {};
      \factor[left=.3 of w1, yshift=+1cm,color=brown] {} {} {w1,b1,q1} {};

      \factor[right=.3 of a2, yshift=-0.5cm,color=brown] {} {} {a2,b2,w2} {};
      \factor[right=.3 of b2, yshift=-0.5cm,color=brown] {} {} {b2,q2,w2} {};
      \factor[right=.3 of q2, yshift=-0.1cm,color=brown] {} {} {a2,q2} {};

      \factor[right=.3 of q2, yshift=-0.5cm,color=brown] {} {} {q2,w2} {};
      \factor[left=.3 of w2, yshift=-0.5cm,color=brown] {} {} {a2,w2} {};
      \factor[left=.3 of w2, yshift=+1cm,color=brown] {} {} {w2,b2} {};

      \factor[right=.3 of a3, yshift=-0.5cm,color=brown] {} {} {a3,b3} {};
      \factor[right=.3 of b3, yshift=-0.5cm,color=brown] {} {} {b3,q3,a3} {};
      \factor[right=.3 of q3, yshift=-0.1cm,color=brown] {} {} {a3,q3} {};

      \factor[right=.3 of q3, yshift=-0.5cm,color=brown] {} {} {q3,w3} {};
      \factor[left=.3 of w3, yshift=-0.5cm,color=brown] {} {} {a3,w3} {};
      \factor[left=.3 of w3, yshift=+1cm,color=brown] {} {} {w3,b3,a3} {};

      \factor[left=0 of w0, yshift=-0.8cm,color=gray] {} {} {w0,tauw0} {};
      \factor[right=.2 of w0, yshift=-0.8cm,color=gray] {} {} {tauw0,w1} {};

      \factor[left=0 of w1, yshift=-0.8cm,color=gray] {} {} {w1,tauw1} {};
      \factor[right=.2 of w1, yshift=-0.8cm,color=gray] {} {} {tauw1,w2} {};

      \factor[left=0 of w2, yshift=-0.8cm,color=gray] {} {} {w2,tauw2} {};
      \factor[right=.2 of w2, yshift=-0.8cm,color=gray] {} {} {tauw2,w3} {};

      \factor[right=-.1 of a0, yshift=+0.8cm,color=gray] {} {} {a0,taua0} {};
      \factor[right=.4 of a0, yshift=+0.8cm,color=gray] {} {} {taua0,a1} {};

      \factor[right=-.1 of a1, yshift=+0.7cm,color=gray] {} {} {a1,taua1} {};
      \factor[right=.4 of a1, yshift=+0.8cm,color=gray] {} {} {taua1,a2} {};

      \factor[right=-.1 of a2, yshift=+0.8cm,color=gray] {} {} {a2,taua2} {};
      \factor[right=.4 of a2, yshift=+0.8cm,color=gray] {} {} {taua2,a3} {};

  \end{tikzpicture}
  \caption{Graph-NLP of the example domain in Sec.~\ref{sec:example}.
    Circles are variables and squares are constraints. We display variables for all mode-switch configurations $(a,b,q,w)$, and trajectory variables ($\tau^a,\tau^w$) (omitting
$\tau^b,\tau^q$ and factors that represent collisions between trajectories to keep the illustration clean). Brown squares are collision avoidance constraints. Gray squares are boundary constraints between trajectories and mode-switches. This representation is similar to the constraint graphs in \cite{garrett2018sampling}.}
  \label{fig:cg_example2}
\end{figure}
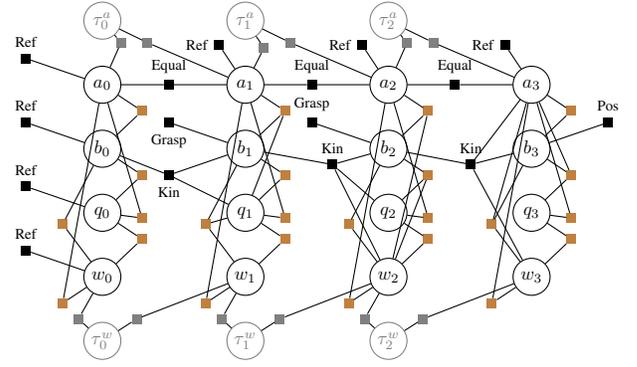

Consider a basic robot manipulation domain, where we have two movable objects, \textit{OA} and \textit{OB}, initially on \textit{OA\_init}, \textit{OB\_init}, and two robot manipulators, \textit{RQ} and \textit{RW}, on a \textit{Table} with the goal to stack \textit{OA} on top of \textit{OB}. 

The logical description contains abstract information of the structure of the configuration (i.e. parent-child relations in the kinematic tree)
but without defining the continuous relative pose. For example, the logic variable \textit{parent\_OA} with domain \textit{\{OA\_init, Table, RQ, RW, OB\}}  indicates whether object \textit{OA} is on the initial position, the \textit{Table}, held by one of the robots or on top of object \textit{OB}. 
The possible logic decision sequences are defined using a PDDL planning task with two action operators: \textit{pick} and \textit{place}.


Figure~\ref{fig:cg_example2} shows the graph-NLP that results from applying the logical decision sequence \textit{pick(OB, OB\_init, RQ)}, \textit{pick(OB, RQ, RW)}, \textit{place(OB, RW, OA)}
from the starting logic state \{\textit{parent\_OA=OA\_init, parent\_OB=OB\_init, gripper\_RQ=free, gripper\_RW=free}\}.

In each vertical slice, we have continuous variables $\{ a , b , q , w , \tau^a , \tau^b , \tau^q , \tau^w \}$, where $a,b$ are the pose of the object with respect to the parent frame in the kinematic chain (for example, using quaternions for the rotation $a,b \in \RR^7$) and $q,w$ are the robot joint configurations (using a 7-DOF manipulator $q,w \in \RR^7$). These variables represent the configurations at the beginning of each motion-phase and are usually called mode-switches.
$ \tau^a , \tau^b , \tau^q , \tau^w \in \RR^{20 \cdot  7}$ are the corresponding trajectories during each motion phase (represented with 20 waypoints). 
These variables are constrained by nonlinear functions that model grasping (\textit{Grasp}), collision avoidance, kinematic switches (\textit{Kin}), position (\textit{Pos}), time-consistency (\textit{equal}) and reference (\textit{Ref}) constraints.

Constraints operate on pairs of consecutive continuous variables, and the constraints that are applied depend on the values of the logical variables.
For instance, a transition of logic variables \textit{Parent\_OA = OA\_init} $\to$ \textit{Parent\_OA = OA\_init}  generates $ ~ a = a'$ (\textit{equal($a,a'$)}).   \textit{Parent\_OB = RQ } 
generates  \textit{grasp(b)}, which constrains the relative position of the object to be inside the two fingers of the gripper with a correct orientation.
  Kinematic switch constraints appear when robots pick/place the objects, e.g., 
  a transition  \textit{Parent\_OB = OB\_init} $\to$ \textit{Parent\_OB' = RQ}  generates \textit{Kin($b,b',q'$)} where \textit{Kin} ensures that the absolute pose of \textit{OB} is kept constant when the robot \textit{RQ} picks the object.

\section{Overview: Graph-NLP Planner}
\label{sec:overview}



\begin{figure}
\begin{tikzpicture}[node distance=1.5cm,
    every node/.style={fill=white, font=\small}, align=center]
  \node (input)     [inout]          {Input: $\langle \mathcal{V}, \mathcal{A},s_0,g, \Pi, \mathcal{H},  \mathcal{X} \rangle$ };
  \node (planner)     [process,below of=input,yshift=.55cm]          {1) Logic planner};

  \node (task)   [left of=planner,xshift=-1.5cm]          {Planning task \\
      $\langle \mathcal{A}', \mathcal{V}', s_0', g' \rangle$

    };


  \node (plan)     [right of=planner,xshift=1.5cm]          {plan\\ $\langle s_0\ldots s_n \rangle$};

  \node (formulation)     [process,below of=plan,yshift=.2cm]          {2) Mapping $\Pi$   };

  \node (graphlgp)     [below of=formulation]          {Graph-NLP \\ $ G( \langle s_0 \ldots s_n \rangle)$ };

  \node (buu)   [data, below of=planner,yshift=+0.4cm, xshift=.2cm]          {};
  \node (buu)   [data, below of=planner,yshift=+0.3cm, xshift=.1cm]          {};

  \node (dataFeas)   [process,below of=planner, yshift=.2cm]          {feasible  \\subgraphs};

  \node (motionplanner)   [process, below of=dataFeas]          {3) Nonlinear solver\\ + conflict extraction };

  \node (reformulation)   [process, below of=task, yshift=.2cm ]          {4) Reformulation};

  \node (ref_feas)   [sec, below of=dataFeas, yshift=+.8cm, xshift=.8cm]          {Sec. \ref{sec:memory}};

  \node (ref_form)   [sec, below of=formulation, yshift=+.8cm, xshift=.7cm]          {Sec. \ref{sec:problem-formulation}};

  \node (output)   [inout, below of=motionplanner,yshift=+.4cm]   {Output: $\langle  (s_0,\xs_0) \ldots  (s_n,\xs_n)  \rangle $ };

  \draw[-]             (planner) -- (plan);

  \node (ref_mot)   [sec, below of= motionplanner, yshift=+.85cm, xshift=.8cm]          {Sec. \ref{sec:find-minimal}};


  \node (buu)   [data, below of=reformulation,yshift=-0.2cm, xshift=-.2cm]          {};
  \node (buu)   [data, below of=reformulation,yshift=-0.1cm, xshift=-.1cm]          {};

  \node (dataInfeas)   [process,below of=reformulation]          {infeasible \\ subgraphs };


  \draw[->]             (motionplanner) -- (output);

  \draw[<->]             (motionplanner) -- (dataFeas);
  \draw[->]             (motionplanner) -- (dataInfeas);
  \draw[->]             (input) -- ++(-3.2,0) -- ++(0,-.5)  ;
  \draw[-]             (plan) -- (formulation);

  \draw[->]             (formulation) -- (graphlgp);
  \draw[->]             (graphlgp) -- (motionplanner);


  \node (ref_sec)   [sec, below of= reformulation, yshift=+.8cm, xshift=.7cm]          {Sec. \ref{sec:reformulate}};



\draw[->]              (dataInfeas) -- (reformulation);

   \draw[-]             (reformulation) -- (task);
   \draw[->]             (task) -- (planner);
  \end{tikzpicture}
  \caption{Overview of the Graph-NLP Planner for solving a PNTC problem  $\langle \mathcal{V}, \mathcal{A},s_0,g, \Pi,  \mathcal{H} , 
  \mathcal{X}  \rangle$. The solution is a sequence of logic and continuous states  $\langle  (s_0,\xs_0) \ldots  (s_n,\xs_n)  \rangle $. See Sec. \ref{sec:overview}.
}
  \label{fig:flowchart}
\end{figure}
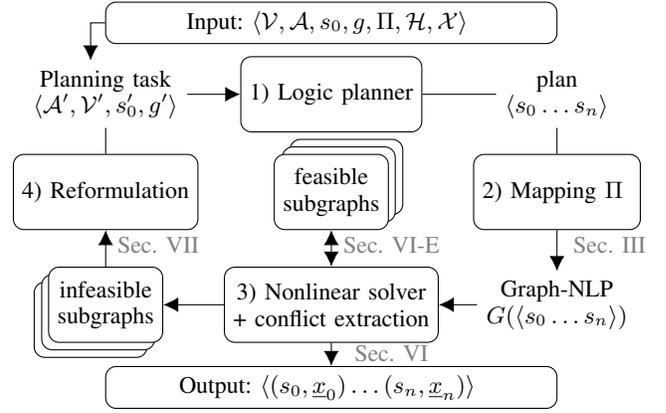

Fig.~\ref{fig:flowchart} provides an overview of the Graph-NLP Planner (GNPP) for solving a PNTC that we will introduce in the next sections. To simplify the presentation, we briefly outline the steps of the algorithm, which are run iteratively:
\begin{enumerate}
  
  \item
  We leverage 
    a state-of-the-art PDDL planner to find a sequence  of logical states that is logically feasible for the current logical planning task.
  \item 
    We generate a nonlinear program with an explicit graph structure, called graph-NLP (Sec. \ref{sec:graphnlp}),
      that represents the continuous variables and the nonlinear constraints defined by the logical state sequence.
  \item 
    An NLP solver evaluates the graph-NLP. If this NLP is feasible, the algorithm terminates and the output is a solution containing all logic and continuous states.
    Otherwise, a minimal conflict in the form of a minimal infeasible subgraph-NLP is extracted and all evaluated subgraph-NLPs are stored in a database of feasible or infeasible subgraphs.
  \item 
    Finally, we reformulate the logical planning task  to  forbid all plans that would generate a graph-NLP that contains any subgraph that was already found to be infeasible. 
\end{enumerate}

\section{Finding small infeasible subgraphs}
\label{sec:find-minimal}



\new{
  In this section, we discuss how to detect a minimal subset of infeasible constraints
from a graph-NLP (Step 3 of the Graph-NLP Planner, Fig. \ref{fig:flowchart}) 
} In the worst case, finding an infeasible subgraph of \emph{minimum cardinality} requires solving a NLP for each subset of constraints $O(2^{|H_G|})$ \cite{shoukry2018smc}. Conversely, a \emph{minimal} infeasible subgraph can be found by solving a linear number of problems  \cite{amaldi1999some}. This search can be accelerated with a divide and conquer strategy, with complexity $O(\log |H_G|)$  \cite{junker2004preferred}.
Recently \cite{shoukry2018smc} presented a technique for finding an approximately minimal subgraph in a convex optimization problem by solving one convex program with slack variables. 

Inspired by these works, we propose an algorithm for finding small minimal infeasible subgraphs that exploits the particular structure of the graph in our setting, namely the time structure of graph-NLPs  and the semantic information contained in them, as well as the convergence point of the nonlinear optimizer. 

\subsection{Double binary search on the time index}

The first key insight is to exploit the time connectivity of our graph-NLP (Property \ref{property:localtime}).
Given an infeasible graph-NLP $G(\langle s_0 \ldots s_n \rangle)$ we can find a minimal temporal subsequence $\langle s_f \ldots s_l \rangle, ~  0 \le f \le l \le n$ such that $G(\langle s_f \ldots s_l \rangle)$  is infeasible with a double binary search that executes $O(\log n)$ calls to a nonlinear optimizer (Properties \ref{propo:minimal-connected} and \ref{propo:monotone}). 
Specifically, we first compute the minimum upper index $l$ such that $G(\langle s_0 \ldots s_l \rangle)$ is infeasible. After fixing $l$ we compute the maximum lower index $f$ such that  $G(\langle s_f \ldots s_l \rangle)$ is infeasible.

\subsection{Relaxations}
\label{sec:relax}


Binary search on time exploits the local connectivity in the temporal dimension, but does not detect the infeasible factors inside an infeasible temporal subsequence.
To address this issue, we propose to solve a set of relaxations of the graph-NLP that evaluates only a subset of variables and constraints. 
Each relaxation corresponds to a subgraph-NLP and is, therefore, a necessary condition of feasibility. The algorithm stores the infeasible relaxations as candidates for the minimal subgraph.



The relaxations depend on the semantic information of the variables and constraints and are problem independent but domain specific.
Intuitively, we are looking for relaxations that make the graph sparser, smaller and potentially disconnected, while keeping those constraints that define the infeasible subgraph. Section \ref{sec:relax_tamp} presents informative relaxations in the context of Task and Motion Planning.  

\subsection{Leveraging the convergence point of the optimizer}

A powerful heuristic to discover a smaller infeasible subset of variables and constraints is to check the convergence point of the optimizer in an infeasible graph.

Given a graph-NLP $G = ( X_G \cup H_G , E_G ) $, the nonlinear optimizer aims to compute $x_i^k ~ \text{s.t} ~ h(x) \le \bm{0} ~ \forall h \in H_G$.
Typical optimization methods converge also for infeasible $G$, where we can use the convergence point $x^*$ as a heuristic guess to find a subgraph of $G$ that is infeasible.
Specifically, we test the subgraph spanned by the constraints violated at $x^*$, i.e $M' = ( X' \cup H' , E' ) $ where $H' = \{ h \in H_G : h(x^*)>0 \}$, $ X' = \{ x \in X_G : \exists h \in \text{Neigh}(x) ~ \text{s.t} ~  h \in H' \}$.
If $M'$ is also infeasible, we consider only $M'$ as a candidate for the minimal infeasible subgraph.




\subsection{The complete algorithm}






\new{
We combine these three ideas into an algorithm to find an infeasible subgraph (Alg. 
1
of  Appendix 
A
in the extended version of the paper\footnote{Available in the project webpage \link{https://quimortiz.github.io/graphnlp/}}).
}
In this algorithm, we alternate between applying relaxations (each relaxation considers only a subset of variables and constraints)
that potentially break the full problem into disconnected components, and computing the minimal infeasible time slice inside each connected component (with double binary search). The convergence point of the optimizer is used to reduce the size of the output infeasible subgraph. 
The algorithm will return the first infeasible subgraph it finds, and therefore it is best to try the relaxations in a \textit{loose} to \textit{tight} order, as this will likely result in a smaller infeasible subgraph.

Deciding whether a relaxation should be applied before or after the binary search on the time index is rather arbitrary.
To this end, a relevant observation is that solving a small NLP that is feasible is usually an order of magnitude faster than checking that a larger NLP is infeasible. Thus, we try to solve numerous small and feasible problems first.








\subsection{Database of feasible subgraphs} 

\label{sec:memory}

The graph structure of the graph-NLP is a suitable representation to share information about feasibility between different sequences of logical states. 
Graphs of different symbolic plans contain common subgraphs, which correspond to sequences of partial states that appear in both plans (potentially at different time indices).


During the execution of the graph-NLP Planner (see Fig. \ref{fig:flowchart}), all solved subgraphs are stored either in a feasible or infeasible database.
Before solving a  new nonlinear program, we check if it is a subgraph of any graph in the feasible database. This check requires a graph isomorphism  test \cite{cordella2004sub}, based on the adjacency structure and semantic information of variable-vertices, which corresponds to the variable-index $k=1 \ldots K$ (without considering the time index) and the name of the constraint $h \in \mathcal{H}$. 
Given the available semantic information, the test is fast in practice (with complexity closer to $O((Kn)^2)$ instead of the worst case exponential).

\subsection{Infeasible subgraphs in TAMP}
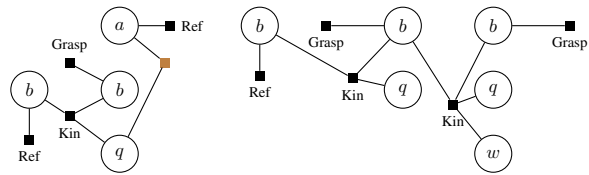
\begin{figure}
  \centering

  \begin{tabular}{cc}

\begin{tikzpicture}[scale=0.7,every node/.style={transform shape}]
        \node[latent] (b0) {$b$} ;
        \node[latent,right=1 of b0  ] (b1) {$b$} ;
        \node[latent,above=.5 of b1] (a1) {$a$} ;
        \node[latent,below=.5 of b1] (q1) {$q$} ;
      \factor[left=.5 of b1, yshift=0.5cm] {trajp0} {above:Grasp} {b1} {};
      \factor[left=.5 of b1, yshift=-.5cm] {trajp0} {below:Kin} {b0, q1,b1} {};
      \factor[below=1 of b0, yshift=0.5cm] {trajp0} {below:Ref} {b0} {};
      \factor[right=.4 of a1,yshift=-.7cm,color=brown] {} {} {a1,q1} {};
      \factor[right=.5 of a1] {trajp0} { right:Ref } {a1} {};
    \end{tikzpicture}&

\begin{tikzpicture}[scale=0.7,every node/.style={transform shape}]
        \node[latent] (b0) {$b$} ;

        \node[latent,right=2 of b0] (b1) {$b$} ;
        \node[latent,below=.5 of b1] (q1) {$q$} ;

        \node[latent,right=1 of b1] (b2) {$b$} ;
        \node[latent,below=.5 of b2] (q2) {$q$} ;
        \node[latent,below=.5 of q2] (w2) {$w$} ;

      \factor[below=1 of b0, yshift=0.5cm] {trajp0} {below:Ref} {b0} {};
      \factor[left=1 of b1 ] {trajp0} {below:Grasp} {b1} {};
      \factor[right=1 of b2] {trajp0} {below:Grasp} {b2} {};

      \factor[left=.5 of b1, yshift=-1cm] {trajp0} {below:Kin} {b0, q1,b1} {};
      \factor[right=.5 of b1, yshift=-1.5cm] {trajp0} {below:Kin} {b1, q2,b2,w2} {};
  \end{tikzpicture}

  \end{tabular}

  \caption{
  Two examples of possible infeasible subgraphs of the graph-NLP of the example domain (Fig. \ref{fig:cg_example2}). \textit{Left}: the robot \textit{RQ} can not pick object \textit{OB} from the initial position  if \textit{OA}  is on the initial position, e.g.  \textit{OA} blocks the grasp of  \textit{OB}. \textit{Right}: it is not possible to pick object \textit{OB} with the robot \textit{RQ} and then do a handover to robot \textit{RW}, e.g. due to kinematic constraints, robot $RQ$ can only pick the object in a certain way that prevents doing a handover later.}
    \label{fig:subgraphs}
\end{figure}

To conclude the section, Fig. \ref{fig:subgraphs}  provides two examples of possible infeasible subgraphs of the graph-NLP  of the example domain (Fig. \ref{fig:cg_example2}), together with an intuitive explanation of the underlying reason of continuous infeasibility. 

\section{Logic Reformulation}

\label{sec:reformulate}

\new{
In this section we discuss how to reformulate the logic task with the information about the continuous infeasibilty (Step 4 of the Graph-NLP planner, Fig. \ref{fig:flowchart})}. Specifically,
given an infeasible subgraph $M$, we modify the logical planning task $\langle \mathcal{V},\mathcal{A},s_0,g\rangle$, to ensure that the logical planner will never generate plans whose graph-NLP contains $M$.
The mapping is achieved through a two-step process:





First, we translate the infeasible subgraph $M=(X_M \cup H_M, E_M)$ into a sequence of logical partial states. Recall that each constraint $h\in H_M$ was generated by
the forward mapping $\Pi(p,p') \mapsto h$. We now trace this mapping back, to get $(p,p')$ which generated $h$.
%
The relative temporal order of the partial states is kept, resulting in the  sequence $\langle  p_0 \ldots p_L \rangle$

Given an infeasible sequence of partial states \seq{p_0}{p_L}, we introduce a compilation that eliminates plans that contain \seq{p_0}{p_L} starting at any time index, similarly to the plan forbidding compilation \cite{katz2018novel}. Our compilation introduces binary symbolic variables $b_l=\{0,1\}, l=0 \ldots L$ to indicate whether the path from $s_0$ to  $s_n$ contains the infeasible subsequence of partial states. $b_l=1$ means that the current path contains the first $l+1$ elements of the infeasible sequence. 


Given a planning task $ \langle \mathcal{V}, \mathcal{A},s_0,g \rangle$ and an infeasible sequence \seq{p_0}{p_L},
the new SAS+ task is $ \langle \mathcal{V'}, \mathcal{A'},s_0',g' \rangle $, where:
\begin{itemize}
  \item $\mathcal{V'} = \mathcal{V} \cup \{ b_0,\ldots,b_L \}$
  \item $s_0'= s \cup \{ b_l=0 \mid l=1,\ldots,L \} \cup \{ b_0 = 1 ~ \text{if} ~ p_0 \subseteq s_0  ; ~ b_0=0 ~ \text{otherwise} \}$
  \item $g'= g \cup \{ b_L=0 \} $
  \item $\mathcal{A'}= \{ a' = \text{mod}(a), ~ a \in \mathcal{A} \}$
\end{itemize}
\noindent
where $a' = \text{mod}(a)$ modifies action $a$ by adding conditional effects to ensure that if action $a$ was executed when $b_{l-1}=1$, and executing $a$ makes $p_l$ true, then $a$ sets $b_l=1$ and $b_{l-1}=0$. Alternatively, if $a$ was executed when $b_{l-1}=1$, and it does not make $p_l$ true, then $a$ sets $b_{l-1}=0$.
The last binary variable $b_{L}$ can not transition $1 \to 0$ (i.e. $b_L=1$ is a dead end).
\new{The formal reformulation $a'= \text{mod}(a)$ is shown in  
  Appendix 
B.}
We can now state the proposition which shows that this compilation eliminates exactly all solutions which satisfy \seq{p_0}{p_L}: 
\begin{proposition}
  \label{thm:forbid}
Let $T = 
\langle \mathcal{V}, \mathcal{A},s_0,g \rangle$ be a SAS+ planning task, \seq{p_0}{p_L} be some infeasible sequence, and $T' = \langle \mathcal{V'}, \mathcal{A'},s_0',g' \rangle$ be the reformulation described above. A plan $\pi$ is a solution of $T'$ iff  $\pi$ is a solution of $T$ and the states along $\pi$ do not contain any subsequence of states $\langle s'_i\ldots s'_{i+L} \rangle$ such that $p_l \subseteq s'_{i+l} $ for $l = 0 \ldots L$ for some $i$.
\end{proposition}

Multiple infeasible subsequences are forbidden by iterative reformulation. 
We are now ready to discuss the properties of our Graph-NLP Planner (Fig. \ref{fig:flowchart}):
\begin{theorem}
  \label{thr:thetheorem}
If the underlying classical planner is sound and complete and the nonlinear optimizer always finds a feasible solution if such exists, then the Graph-NLP Planner is sound and complete.
\end{theorem}

\textit{Proof Sketch:} The proof follows from the fact that any subsequence we forbid can not be part of any feasible solution (because it generates a subgraph found to be infeasible, Prop. \ref{prop:subgraph}), together with the fact that our compilation eliminates only plans which contain these subsequences (Prop. \ref{thm:forbid}).
The completeness of the algorithm does not require the infeasible subgraphs to be minimal, nor the mapping  $\Pi$. Nonetheless, these properties are desirable for an efficient algorithm.

\section{Experimental Results}

\label{sec:experimental_results}

\subsection{Benchmark}

Our algorithm is evaluated in 3 different simulated scenarios, where  the goal is to move obstacles, rearrange and stack up to six blocks to build towers with several robots. The evaluation on real robots is reported in Section \ref{sec:real}.
\begin{enumerate}
  \item \textit{Laboratory (Lab)}: Two 7-DOF manipulator arms execute pick and place actions to build a tower. The solution requires handovers, regrasping and removing obstacles. It is based on the real-world setting, Fig. \ref{fig:real}. 
  \item \textit{Workshop (Work)}: Extension of the \textit{Laboratory} scenario that includes four robots  and a stick, that can be grasped and used as a tool to reach blocks, Fig \ref{fig:showcase}.
  \item \textit{Field}: Contains a fixed 7-DOF manipulator and a mobile 7-DOF manipulator, with two additional actions operators: \textit{start-move} and \textit{end-move}, for moving the base of the mobile robot on the floor, Fig. \ref{fig:showcase}.
\end{enumerate}

For each scenario, we generate 5 different problems (e.g \textit{Lab\_\{1,2,3,4,5\}}) by modifying the symbolic goal and the initial configuration to increase the complexity in the logical and geometric levels, while keeping the number of movable objects constant (except for the
easier versions \textit{Work\_1} and \textit{Field\_1}). A subset of these problems, together with the computed solutions, are shown in the project webpage.




\subsection{Relaxations for finding infeasible subgraphs}
\label{sec:relax_tamp}


The formulation \textit{PNTC} and the solver is general and domain independent. Domain knowledge is introduced through the relaxations used for extracting minimal conflicts (Sec. \ref{sec:relax}). The following relaxations (applicable in any TAMP problem) are used in the benchmark scenarios:

\subsubsection{Removal of trajectories} The remaining graph only contains variables for the mode-switches, considerably reducing the dimensionality of the nonlinear program while still detecting most of the geometric infeasibilities.


\subsubsection{Removal of collision constraints} Collision constraints connect all robot configuration and object pose variables in the same time step, resulting in a densely connected graph (see Fig. \ref{fig:cg_example2}). Without collisions, the graph becomes sparse, and object and robot variables are only connected by grasping, kinematics and placement constraints.

\subsubsection{Removal of time consistency} Time-consistency constraints (\textit{Equal} in Fig. \ref{fig:cg_example2}) appear when objects are not modified by an action. This relaxation does not consider the long term dependencies of the manipulation sequence and creates a sparse time structure.

\subsubsection{Removal of robots variables} The remaining graph considers only the variables for the objects, detecting infeasible placements due to collisions between objects.

\begin{table*}

\caption{Number of NLP evaluations and CPU time, averaged over 10 random seeded runs, with standard deviations in gray. }
\begin{tabular}{lllrrrrrrrrrrrr} 
\toprule 
{} & \multicolumn{2}{c}{\textit{length}} & \multicolumn{2}{c}{\textit{One-way}} & \multicolumn{2}{c}{\textit{MBTS}} & \multicolumn{2}{c}{\textit{GNPP\_t}} & \multicolumn{2}{c}{\textit{GNPP\_tr}} & \multicolumn{2}{c}{\textit{GNPP\_trn}} & \multicolumn{2}{c}{\textit{GNPP\_trng}} \\ 
\cmidrule(lr){2-3}
\cmidrule(lr){4-5}
\cmidrule(lr){6-7}
\cmidrule(lr){8-9}
\cmidrule(lr){10-11}
\cmidrule(lr){12-13}
\cmidrule(lr){14-15}
{} & {$N_0$} & {$N$} & {\textit{NLP}} & {\textit{time}} & {{\textit{NLP}}} & {\textit{time}} & {{\textit{NLP}}} & {\textit{time}} & {\textit{NLP}} & {\textit{time}} & {\textit{NLP}} & {\textit{time}} & {\textit{NLP}} & {\textit{time}} \\ 
\midrule 
Work\_1 & 2 & 4 & 77.0{\scriptsize \color{gray} 0.0}& 8.2{\scriptsize \color{gray} 1.0}& 371{\scriptsize \color{gray} 58.5}& 23.5{\scriptsize \color{gray} 4.4}& 50.8{\scriptsize \color{gray} 12.0}& 5.7{\scriptsize \color{gray} 1.2}& 53.0{\scriptsize \color{gray} 12.6}& 5.7{\scriptsize \color{gray} 1.3}& 60.8{\scriptsize \color{gray} 10.5}& 5.6{\scriptsize \color{gray} 0.9}& 55.2{\scriptsize \color{gray} 1.4}& \textbf{5.3}{\scriptsize \color{gray} 0.1}\\ 
Work\_2 & 4 & 6 & - & - & - & - & - & - & 113{\scriptsize \color{gray} 40.4}& 22.7{\scriptsize \color{gray} 9.2}& 94.7{\scriptsize \color{gray} 1.7}& 19.7{\scriptsize \color{gray} 6.0}& 86.8{\scriptsize \color{gray} 1.5}& \textbf{16.7}{\scriptsize \color{gray} 2.9}\\ 
Work\_3 & 4 & 6 & - & - & - & - & - & - & 105{\scriptsize \color{gray} 37.0}& \textbf{19.1}{\scriptsize \color{gray} 5.3}& 95.6{\scriptsize \color{gray} 1.0}& 22.1{\scriptsize \color{gray} 5.9}& 86.1{\scriptsize \color{gray} 1.5}& 21.4{\scriptsize \color{gray} 6.1}\\ 
Work\_4 & 8 & 10 & - & - & - & - & - & - & 282{\scriptsize \color{gray} 0.0}& \textbf{53.1}{\scriptsize \color{gray} 5.6}& 307{\scriptsize \color{gray} 1.9}& 56.4{\scriptsize \color{gray} 7.6}& 270{\scriptsize \color{gray} 1.8}& 55.4{\scriptsize \color{gray} 9.0}\\ 
Work\_5 & 8 & 11 & - & - & - & - & - & - & - & - & 355{\scriptsize \color{gray} 4.9}& \textbf{76.0}{\scriptsize \color{gray} 7.0}& 309{\scriptsize \color{gray} 5.3}& 76.8{\scriptsize \color{gray} 9.4}\\ 
Lab\_1 & 2 & 3 & 25.0{\scriptsize \color{gray} 0.0}& 7.1{\scriptsize \color{gray} 1.0}& 25.0{\scriptsize \color{gray} 0.0}& 4.1{\scriptsize \color{gray} 0.5}& 21.0{\scriptsize \color{gray} 0.0}& 4.5{\scriptsize \color{gray} 0.2}& 25.0{\scriptsize \color{gray} 0.0}& \textbf{3.2}{\scriptsize \color{gray} 0.2}& 30.0{\scriptsize \color{gray} 0.0}& 3.3{\scriptsize \color{gray} 0.1}& 28.0{\scriptsize \color{gray} 0.0}& 3.3{\scriptsize \color{gray} 0.2}\\ 
Lab\_2 & 2 & 3 & 12.0{\scriptsize \color{gray} 0.0}& 3.1{\scriptsize \color{gray} 0.5}& 28.9{\scriptsize \color{gray} 0.3}& 3.7{\scriptsize \color{gray} 0.5}& 32.0{\scriptsize \color{gray} 0.0}& 5.5{\scriptsize \color{gray} 0.3}& 46.0{\scriptsize \color{gray} 0.0}& 4.3{\scriptsize \color{gray} 0.2}& 23.0{\scriptsize \color{gray} 0.0}& \textbf{2.1}{\scriptsize \color{gray} 0.1}& 21.0{\scriptsize \color{gray} 0.0}& \textbf{2.1}{\scriptsize \color{gray} 0.1}\\ 
Lab\_3 & 4 & 5 & 19.0{\scriptsize \color{gray} 0.0}& 8.4{\scriptsize \color{gray} 1.2}& 34.0{\scriptsize \color{gray} 0.0}& 18.5{\scriptsize \color{gray} 2.7}& 26.0{\scriptsize \color{gray} 0.0}& 5.9{\scriptsize \color{gray} 0.4}& 23.0{\scriptsize \color{gray} 0.0}& \textbf{3.1}{\scriptsize \color{gray} 0.2}& 25.0{\scriptsize \color{gray} 0.0}& 3.3{\scriptsize \color{gray} 0.4}& 24.0{\scriptsize \color{gray} 0.0}& 3.2{\scriptsize \color{gray} 0.2}\\ 
Lab\_4 & 4 & 9 & - & - & - & - & - & - & - & - & 70.0{\scriptsize \color{gray} 0.0}& 6.5{\scriptsize \color{gray} 0.6}& 60.1{\scriptsize \color{gray} 3.5}& \textbf{6.3}{\scriptsize \color{gray} 0.4}\\ 
Lab\_5 & 12 & 17 & - & - & - & - & - & - & 87.0{\scriptsize \color{gray} 0.0}& \textbf{18.7}{\scriptsize \color{gray} 1.8}& 93.0{\scriptsize \color{gray} 0.0}& 19.1{\scriptsize \color{gray} 2.0}& 83.0{\scriptsize \color{gray} 0.0}& 19.0{\scriptsize \color{gray} 2.0}\\ 
Field\_1 & 2 & 4 & 19.0{\scriptsize \color{gray} 0.0}& 7.0{\scriptsize \color{gray} 2.0}& 90.0{\scriptsize \color{gray} 0.0}& 15.3{\scriptsize \color{gray} 3.1}& 19.0{\scriptsize \color{gray} 0.0}& 5.7{\scriptsize \color{gray} 1.1}& 14.1{\scriptsize \color{gray} 0.3}& 2.9{\scriptsize \color{gray} 1.4}& 16.0{\scriptsize \color{gray} 0.0}& \textbf{2.6}{\scriptsize \color{gray} 0.3}& 16.0{\scriptsize \color{gray} 0.0}& 3.1{\scriptsize \color{gray} 1.0}\\ 
Field\_2 & 2 & 6 & - & - & - & - & - & - & 46.0{\scriptsize \color{gray} 0.0}& \textbf{6.1}{\scriptsize \color{gray} 0.4}& 53.0{\scriptsize \color{gray} 0.0}& 6.3{\scriptsize \color{gray} 0.5}& 52.5{\scriptsize \color{gray} 0.5}& 6.3{\scriptsize \color{gray} 0.5}\\ 
Field\_3 & 4 & 8 & - & - & - & - & - & - & 75.0{\scriptsize \color{gray} 0.0}& \textbf{11.7}{\scriptsize \color{gray} 1.2}& 84.0{\scriptsize \color{gray} 0.0}& 12.3{\scriptsize \color{gray} 1.4}& 78.6{\scriptsize \color{gray} 0.5}& \textbf{11.7}{\scriptsize \color{gray} 0.7}\\ 
Field\_4 & 6 & 10 & - & - & - & - & - & - & 67.0{\scriptsize \color{gray} 0.0}& \textbf{13.2}{\scriptsize \color{gray} 1.5}& 77.0{\scriptsize \color{gray} 0.0}& 13.6{\scriptsize \color{gray} 1.6}& 76.0{\scriptsize \color{gray} 0.0}& 13.5{\scriptsize \color{gray} 1.3}\\ 
Field\_5 & 6 & 11 & - & - & - & - & - & - & 282{\scriptsize \color{gray} 0.0}& 56.5{\scriptsize \color{gray} 6.6}& 289{\scriptsize \color{gray} 1.0}& \textbf{50.7}{\scriptsize \color{gray} 5.5}& 262{\scriptsize \color{gray} 0.5}& 51.4{\scriptsize \color{gray} 6.6}\\ 
\bottomrule 
\end{tabular} 

\label{table:thetable}
\end{table*}

\subsection{Algorithms under Comparison}

We compare our approach with two different formulations that combine a logical search with joint nonlinear optimization for solving Task and Motion Planning problems.

\subsubsection{One-way interface between Top-K Planning and a nonlinear optimizer (One-way)}
This baseline combines Top-K planning \cite{katz2018novel} to generate a set of different logic plans with a nonlinear optimizer to evaluate the plans. The planner does not receive any information about the geometric reason of infeasibility and only blocks the evaluated plans.
\subsubsection{Multibound Tree Search (MBTS)}
The MBTS Solver \cite{toussaint2017multi} incrementally builds a tree in a breadth-first order to explore sequences of logic actions that reach the symbolic goal.
\new{
  Instead of solving the full continuous optimization problem directly, MBTS computes first relaxed versions (\textit{bounds})
that consider a subset of variables and constraints. The  \textit{pose bound} optimizes each mode-switch independently and  the \textit{sequence bound} considers the full sequence of mode-switches, that are optimized jointly. 
}




\subsubsection{Four Variations of our Graph-NLP Planner (GNPP)} 
We evaluate our full planner \textit{GNPP\_trng}, and 
three additional versions:  \textit{GNPP\_t}, \textit{GNPP\_tr}, and  \textit{GNPP\_trn} to do an ablation study of the algorithm to extract infeasible subgraphs.
Suffixes indicate: $t$=time search, $r$=relaxation, $n$=convergence heuristic, and $g$=feasible graph database.

\subsection{Metrics}

Each algorithm is run 10 times with different random seeds and a timeout of 100 seconds. For each method we report on the number of solved NLPs (\textit{NLP}) and the CPU time in seconds (\textit{time}) in Table \ref{table:thetable}. ``--'' means failure to find a solution within 100 seconds with at least 70\% success rate.

\textit{Time}\footnote{Experiments are run on Single Core i7-1165G7@2.80GHz} provides an objective way to compare algorithms that use different underlying methods.
The number of solved NLPs is informative but does not capture 
the influence of the size and feasibility of NLPs on the running time of the solver.

For each problem, \textit{N} denotes the length of the shortest found plan that is both logically and geometrically feasible and $N_0$ is the length of the logical plan that solves the initial logical task (that is, without considering the continuous information).
\textit{N} and $N_0$ are a proxy for the difficulty: the number of candidate plans typically grows exponentially with $N$, and the difference $N-N_0$ shows the 
impact
of the continuous domain 
on the logical planner.
The approximate branching factor is 12 in  \textit{Lab\_\{1,2,3,4,5\}},
13 in \textit{Field\_\{2,3,4,5\}}, 24 in \textit{Work\_\{2,3,4,5\}}, 4 in \textit{Work\_1} and 5 in \textit{Field\_1}.


\subsection{Comparison to baselines}

Concerning the problems solved, \textit{One-way} and \textit{MBTS} can only solve the easier problems in each scenario while \textit{GNPP\_trn/trng} solves all the problems. Our algorithm is significantly faster in the problems solved by \textit{One-way} and \textit{MBTS}, where the more efficient encoding of geometric information reduces the running time. 

The success rate of our planners \textit{GNPP\_trn/trng} is 100\% in all problems except for \textit{Field\_5} (80\%), \textit{Work\_4} (95\%) and \textit{Work\_5} (90\%), where the optimizer fails to solve feasible graph-NLPs in a few runs. The performance of \textit{GNPP\_trng} is not affected by the branching factor of the underlying problem and provides good scaling with respect to $N$ and $N-N_0$. The highest computational time corresponds to \textit{Field\_5} and \textit{Work\_5}, that require a long plan and detecting collisions between movable objects. \new{In TAMP, the practical size of the graph-NLPs is $O(K^2 n)$ (where $n$ is the length of the action sequence, and $K$ is the number of objects and robots). The domains are modelled using a small set ($<20$) of different types of nonlinear constraints (e.g. Fig. 
\ref{fig:cg_example2}). }




\subsection{Ablation study}






\subsubsection{Analysis of the relaxations}
 \textit{GNPP\_t} detects conflicts of the form \seq{s_i}{s_{i+l}}, while \textit{GNPP\_tr} checks relaxations to generate smaller conflicts \seq{p_i}{p_{i+l}}. Small conflicts lead to more aggressive pruning of logic plans, and are essential to solve the harder problems (the number of solved problem is 5 vs 13 out of 15).
 \new{An analysis of the impact of each relaxation is shown in Appendix 
   C, where we observe that \textit{Removal of trajectories} and \textit{Removal of collision constraints} are the most informative relaxations.
}





\subsubsection{Analysis of the convergence heuristic} The results show that the convergence heuristic is important in problems that require reasoning about the collisions between movable objects, e.g. when the robot must move one object before placing another to avoid a collision. In this case, the relaxations are not informative, while the convergence point of the optimizer in these infeasible problems usually indicates which are the objects that are in collision. 
\textit{GNPP\_tr} solves 13 out of 15 and  \textit{GNPP\_trn} solves 15 out of 15.

\subsubsection{Analysis of the database of feasible graphs}
 \textit{GNPP\_trng} reduces the number of solved NLPs, from a total average of 1673 to 1508, but there is no improvement in the computational time. 
We conjecture that the database approach will provide higher benefits in a setting where solving the NLPs requires more time. 

\subsection{Scalability and limitations}
\new
{
  We conduct two additional experiments in the \textit{Laboratory} scenario to explicitly evaluate the scalability of the method when increasing the number of blocks to be stacked (from 4 to 32) and the number of movable obstacles in a cluttered table (from 1 to 6).
  Results are shown in Appendix 
  D.
  Our planner requires 5.5, 36.6, and 241.9 seconds to compute a plan that stacks, respectively, 8, 16 and 28 blocks in pairs; and 11.8, 22.3 and 54.7 seconds to rearrange 6 blocks while clearing first 1, 3 and 6 obstacles.



  The running time of the Graph-NLP Planner scales polynomially with the number of objects and plan length, and the 
practical bottleneck is the time spent on solving large nonlinear programs (with cubic complexity on the number of objects and linear on the plan length). }
\new{
The main weakness of our method is that the nonlinear optimizer is not guaranteed to find a solution for a (sub)graph-NLP even if one exists, given that the nonlinear constraints define a non-convex optimization problem (which could break the assumption in Thm. \ref{thr:thetheorem}). However, the extensive experiments 
demonstrate that the solver is efficient and reliable in relevant use-cases of TAMP.
}

\begin{figure}
  \begingroup
\setlength{\tabcolsep}{2pt} 

  \begin{tabular}{ccc}

    \includegraphics[width=.32\linewidth]{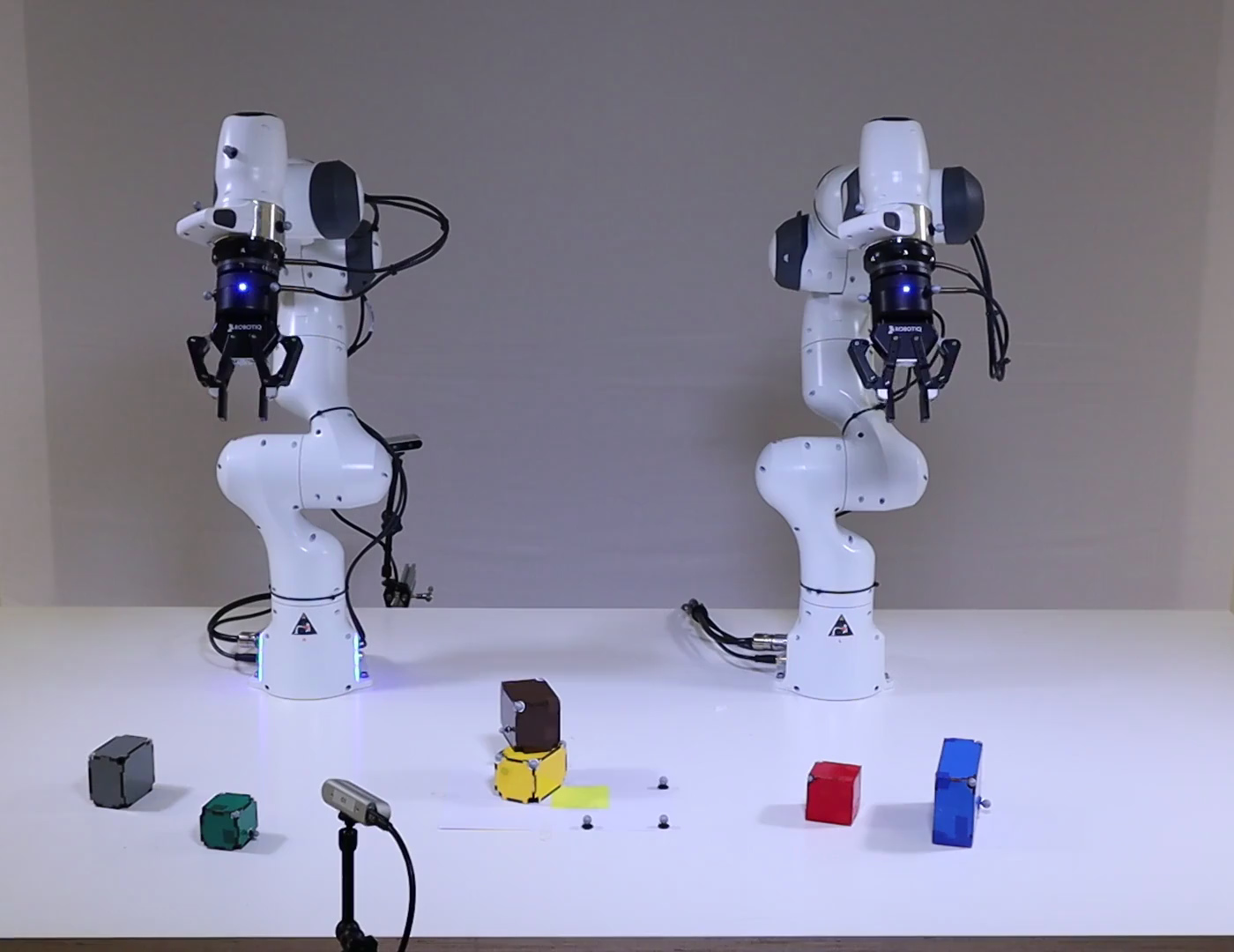} &
    \includegraphics[width=.32\linewidth]{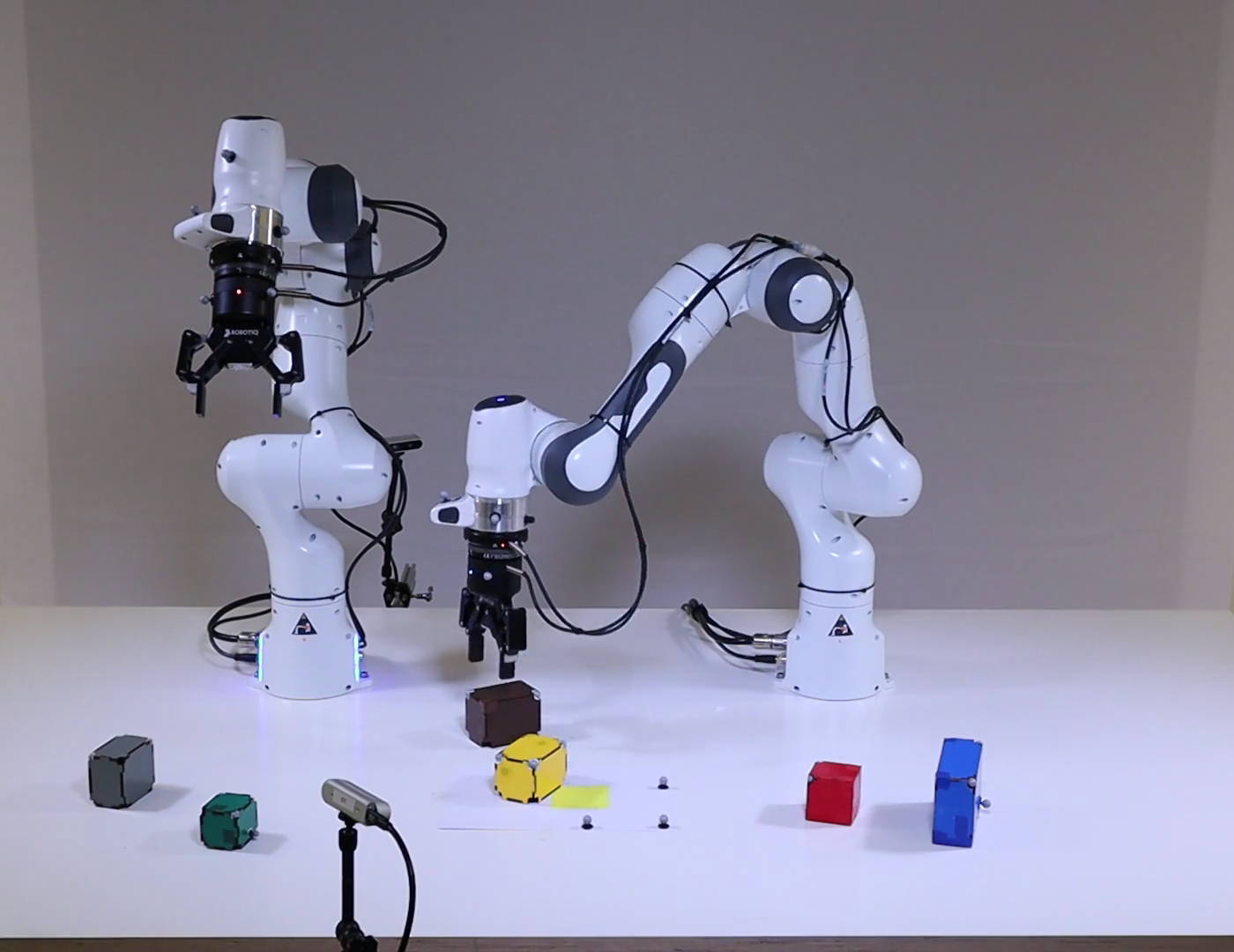} &
    \includegraphics[width=.32\linewidth]{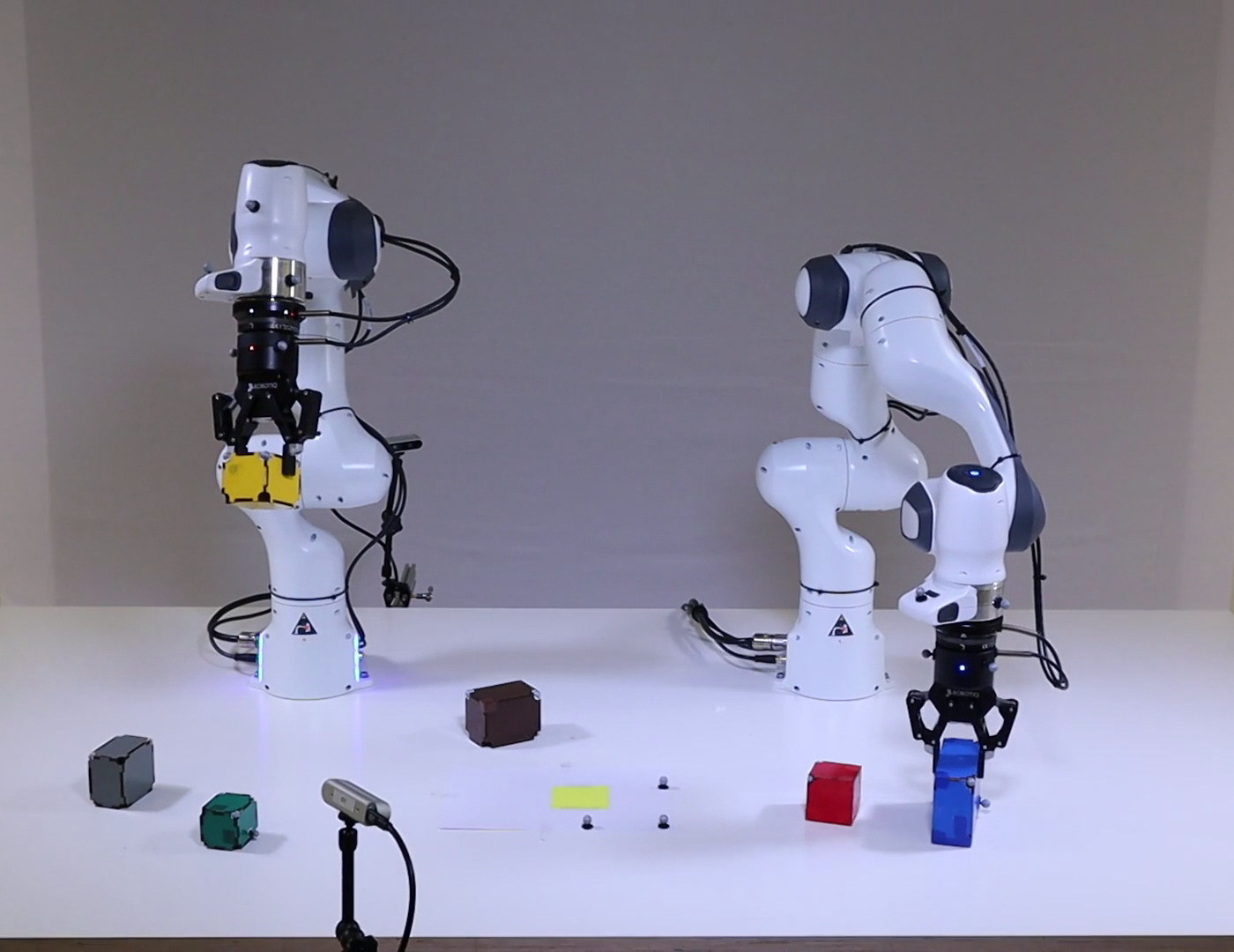} \\

    \includegraphics[width=.32\linewidth]{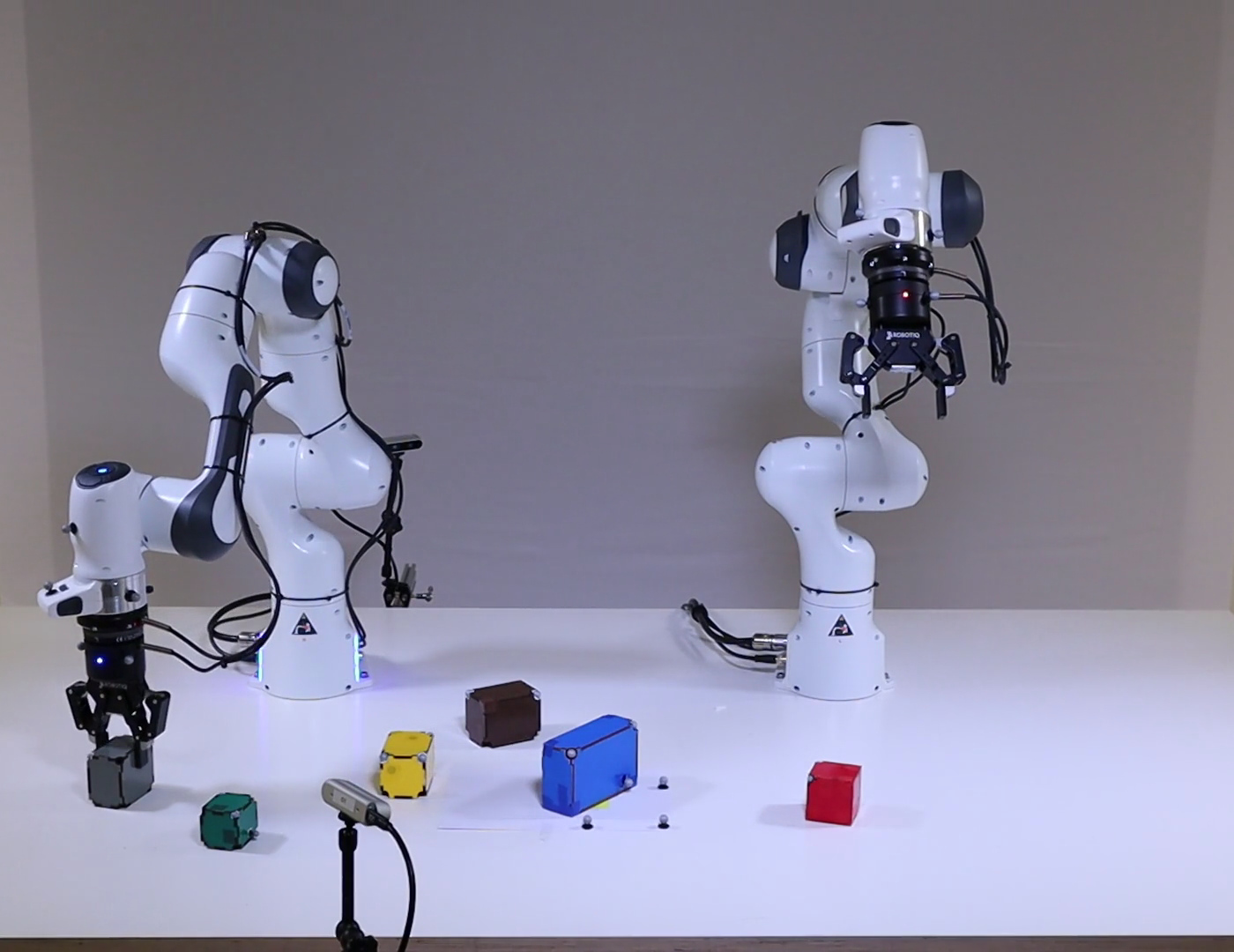} &
    \includegraphics[width=.32\linewidth]{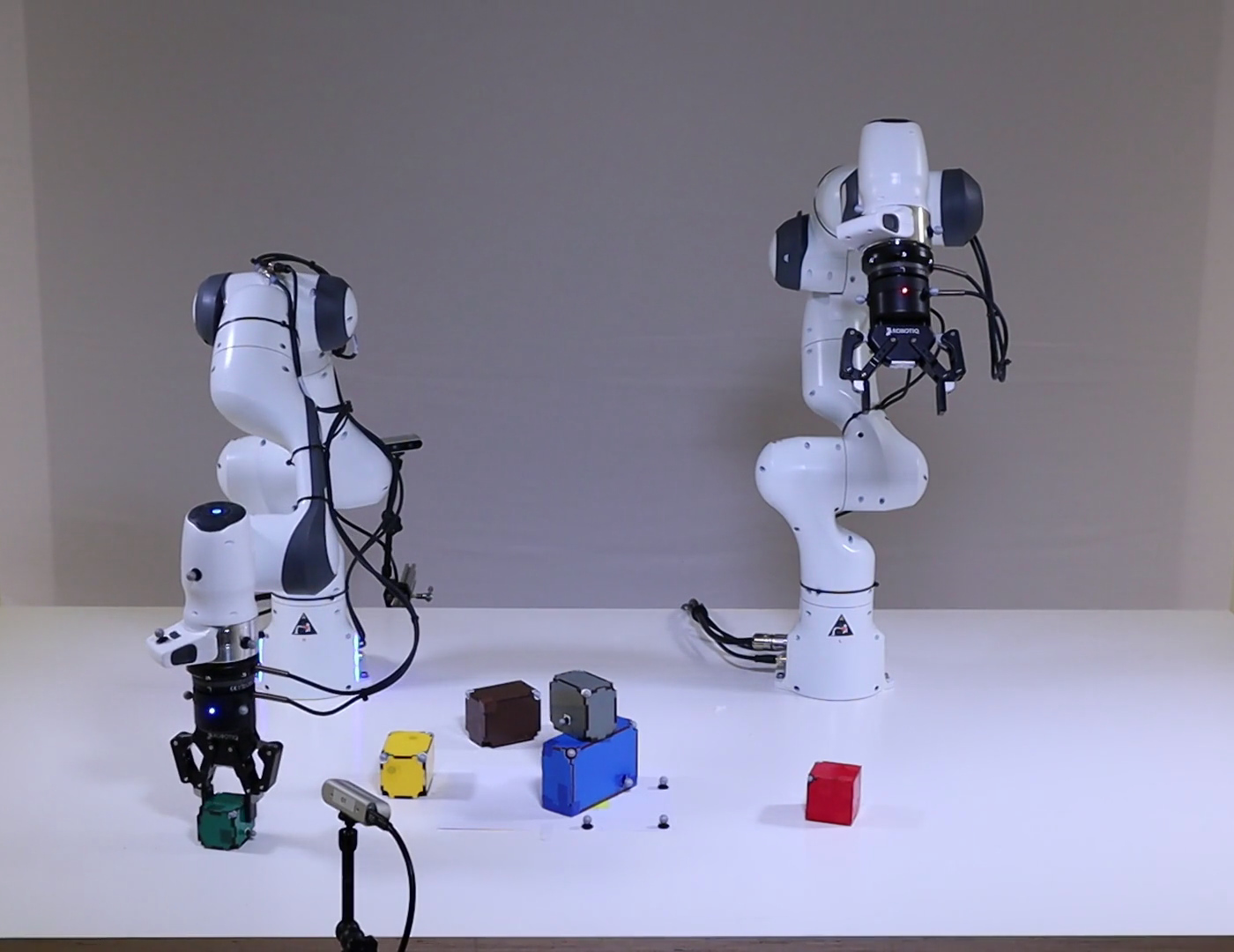} &
    \includegraphics[width=.32\linewidth]{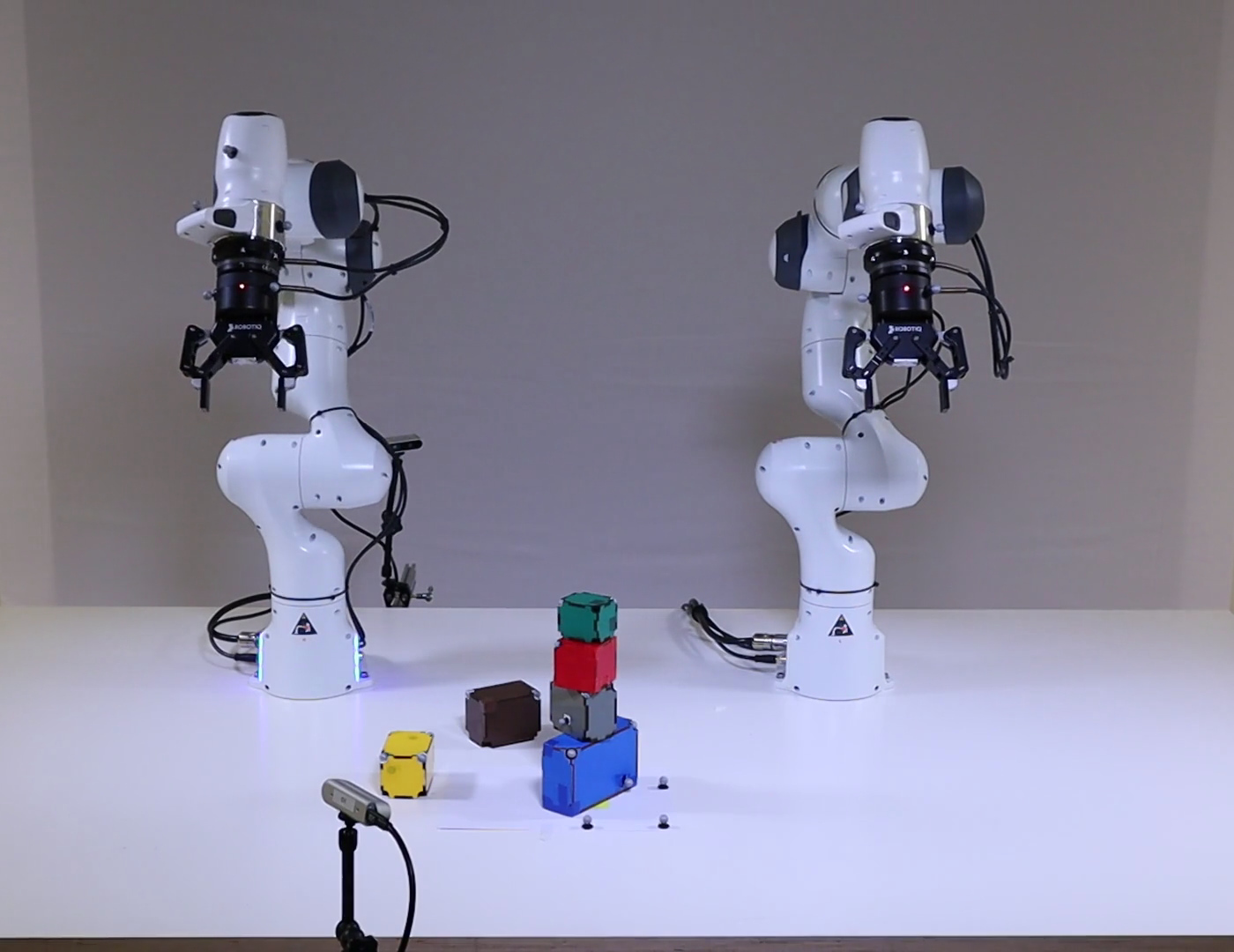}



  \end{tabular}
  \endgroup
  \caption{The task in this real-world experiment is to build the tower \textit{blue-gray-red-green} in the central spot (highlighted in yellow). The solution,  computed in only 8.88 seconds with our planner, requires moving the \textit{brown} and \textit{yellow} blocks to avoid collisions and executing a total of 12 actions.}
  \label{fig:real}
\end{figure}

\subsection{Real-time planning in the real-world}
\label{sec:real}
We demonstrate our solver in a real-world version of the \textit{Laboratory} environment (two 7-DOF Manipulators and up to 6 movable objects), see Fig. \ref{fig:real}.
%
The solver is integrated in a \textit{Sense}-\textit{Plan}-\textit{Act} pipeline, where we first perceive the scene with an external motion capture system, compute a full (logical and continuous) plan and execute the plan. 





The real world evaluation consists of three  scenarios: \textit{Tower}, \textit{Hanoi-Tower}, \textit{Obstacles-Tower}, for a total of 11 problems.
The symbolic goal is to build a tower of cubes at different locations: \textit{Hanoi-Tower} introduces the classical Hanoi logic constraints 
and \textit{Obstacles-Tower} requires plans that clear obstacles. 
The planning time differs across problems: 2.8 s to build a tower of 6 blocks in the center of the table (12 actions), 
8.8 s to remove two obstructing blocks and stack 4 blocks (12 actions), 9.4 s to build a Hanoi Tower (12 actions) and 27.2 s to remove obstructing blocks and  transfer blocks from the left to right side (16 actions). 
Recordings of planning and execution are shown in the project webpage.






\section{Conclusion} 

 We presented a solver that combines nonlinear optimization and PDDL planning for joint optimization of logical and continuous variables in robotic planning.
 The key contribution is the novel bidirectional interface between logic and continuous constraints, realized through the detection of infeasible subgraphs
 and a reformulation to inform the logical planner about subgraph infeasibility.
 The problem formulation is formalized  as \textit{PNTC}, which extends classical planning with nonlinear transition constraints. 

 Our experiments in Task and Motion Planning show that the algorithm is faster and more scalable than Multi-Bound Tree search for LGP, while maintaining the generality and using the same input information.
 These results are further validated in real-world experiments, where our solver generates plans for two 7-DOF robots with 6 objects in few seconds.
As future work, we would like to combine nonlinear optimization with conditional constraint sampling for solving large graph-NLPs, potentially bridging the gap between sample- and optimization-based approaches to TAMP.



\newpage

\bibliography{IEEEabrv,IEEEexample}




\clearpage
\appendix


\subsection{Algorithm for finding infeasible subgraphs}
\label{app:algo}

Algorithm \ref{alg:infeas} shows an heuristic way to combine our three algorithmic ideas 
(``time binary search", ``relaxations" and ``convergence heuristic") to detect small infeasible subgraphs (see Sec. \ref{sec:find-minimal}). It achieves a good experimental performance with an easy implementation. More complex and efficient conflict extraction strategies will be explored in future work.

\begin{algorithm}
\caption{Finding a small infeasible subgraph}\label{alg:infeas}
\begin{algorithmic}
\State \textbf{Input:} Graph-NLP $G( \seq{s_0}{s_n} )$,  set of relaxation rules $R=\{r_j\}$.
\State \textbf{Output:} Small infeasible subgraph $M \subseteq G $.
\For{$r$ in $R$}
\State $G_r \gets r(G)$ \Comment{remove variables, constraints. $G_r \subseteq G$}
\State $\{g_i\} \gets \textbf{Connected\_Components}(G_r)$
\For{$g$ in $\{g_i\}$}
\State feasible, $[t_f, t_l]$ $ \gets $ \textbf{Binary\_Search\_Time}($g$) \\
\Comment{$[t_f, t_l]$  is infeasible time interval (if any)}
\If {not feasible} 
\State $g_s \gets $ \textbf{Convergence\_Heuristic}$(g([t_f,t_l]))$ \\
\Comment{$g([t_f, t_l]) \subseteq g $ only has variables in time \\ 
\hspace{3cm} interval $[t_f, t_l]$ }

\State \textbf{return} $g_s$
\EndIf
\EndFor
\EndFor
\end{algorithmic}
\end{algorithm}

\subsection{Logic reformulation}
\label{sec:formal_reformulation}

Given a planning task $ \langle \mathcal{V}, \mathcal{A},s_0,g \rangle$ and an infeasible sequence of partial states \seq{p_0}{p_L},
the new SAS+ task is $ \langle \mathcal{V'}, \mathcal{A'},s_0',g' \rangle $, where:
\begin{itemize}
  \item $\mathcal{V'} = \mathcal{V} \cup \{ b_0,\ldots,b_L \}$
  \item $s_0'= s \cup \{ b_l=0 \mid l=1,\ldots,L \} \cup \{ b_0 = 1 ~ \text{if} ~ p_0 \subseteq s_0  ; ~ b_0=0 ~ \text{otherwise} \}$
  \item $g'= g \cup \{ b_L=0 \} $
  \item $\mathcal{A'}= \{ a' = \text{mod}(a), ~ a \in \mathcal{A} \}$
\end{itemize}
\noindent

To formally describe $a' = \text{mod}(a)$,
we treat the partial assignment $p_l$ as a set of facts, and add the following conditional effects to $a$:
 $(\bigwedge_{f \in p_0 \setminus \text{eff}(a)} f) \triangleright (b_0 \to 1)$; and for every $l = 1, \ldots, L$: $(b_{l-1} = 1) \wedge (\bigwedge_{f \in p_l \setminus \text{eff}(a)} f) \triangleright (b_{l-1},b_l) \to (0,1)$ and   $(b_{l-1} = 1) \wedge \neg (\bigwedge_{f \in p_l \setminus \text{eff}(a)} f) \triangleright (b_{l-1}) \to 0$, where the notation $A \triangleright B \to 1 $ means ``if $A$ , then $B \to 1$". 
 If the effects of $a$ are inconsistent with $p_l$ (that is, one of the effects of $a$ assigns a different value to one of the variables in $p_l$), the expression $ (\bigwedge_{f \in p_l \setminus \text{eff}(a)} f) $ always evaluates to false.

Finally, we remark that avoiding a sequence of states which satisfies \seq{p_0}{p_L} can be encoded as a PDDL 3 trajectory constraint.
The  above-mentioned compilation is a special case.

\newpage
\subsection{Experimental study of relaxations}
\label{sec:appen_relax}

\begin{table*}[t]
\centering
\begin{tabular}{lllrrrrrrrr} 
\toprule 
{} & \multicolumn{2}{c}{length} & \multicolumn{2}{c}{
\textit{GNPP\_tnr}}
 & \multicolumn{2}{c}
 {
\textit{GNPP\_tnr1}} & \multicolumn{2}{c}{
\textit{GNPP\_tnr2 }} & \multicolumn{2}{c}{GNPP\_tnr3} \\ 
{} & {$N\_0$} & {$N$} & {\textit{NLP}} & {\textit{time}} & {\textit{NLP}} & {\textit{time}} & {\textit{NLP}} & {\textit{time}} & {\textit{NLP}} & {\textit{time}} \\ 
\midrule 
Work\_1 & 2 & 4 & 57.3{\scriptsize \color{gray} 1.9}& 5.3{\scriptsize \color{gray} 0.1}& 93.6{\scriptsize \color{gray} 2.3}& 4.8{\scriptsize \color{gray} 0.1}& 88.8{\scriptsize \color{gray} 1.4}& 5.3{\scriptsize \color{gray} 0.1}& 46.8{\scriptsize \color{gray} 0.7}& 4.7{\scriptsize \color{gray} 0.0}\\ 
Work\_2 & 4 & 6 & 95.0{\scriptsize \color{gray} 0.0}& 15.6{\scriptsize \color{gray} 3.5}& 195{\scriptsize \color{gray} 0.5}& 33.7{\scriptsize \color{gray} 6.0}& 164{\scriptsize \color{gray} 1.4}& 15.4{\scriptsize \color{gray} 2.6}& 78.5{\scriptsize \color{gray} 1.9}& 14.7{\scriptsize \color{gray} 4.1}\\ 
Work\_3 & 4 & 6 & 96.2{\scriptsize \color{gray} 2.4}& 15.7{\scriptsize \color{gray} 3.5}& 196{\scriptsize \color{gray} 0.5}& 30.2{\scriptsize \color{gray} 2.2}& 164{\scriptsize \color{gray} 0.4}& 16.3{\scriptsize \color{gray} 3.7}& 78.1{\scriptsize \color{gray} 1.6}& 13.0{\scriptsize \color{gray} 3.1}\\ 
Work\_4 & 8 & 10 & 308{\scriptsize \color{gray} 2.4}& 52.5{\scriptsize \color{gray} 4.3}& 1042{\scriptsize \color{gray} 2.8}& 111{\scriptsize \color{gray} 0.8}& 744{\scriptsize \color{gray} 5.1}& 57.1{\scriptsize \color{gray} 3.3}& 535{\scriptsize \color{gray} 1.2}& 38.3{\scriptsize \color{gray} 2.8}\\ 
Work\_5 & 8 & 11 & 353{\scriptsize \color{gray} 0.4}& 71.7{\scriptsize \color{gray} 5.6}& - & - & 874{\scriptsize \color{gray} 24.7}& 79.6{\scriptsize \color{gray} 5.7}& 633{\scriptsize \color{gray} 2.4}& 51.8{\scriptsize \color{gray} 3.6}\\ 
Lab\_1 & 2 & 3 & 30.0{\scriptsize \color{gray} 0.0}& 3.6{\scriptsize \color{gray} 0.1}& 57.0{\scriptsize \color{gray} 0.0}& 4.9{\scriptsize \color{gray} 0.1}& 46.0{\scriptsize \color{gray} 0.0}& 3.6{\scriptsize \color{gray} 0.1}& 30.0{\scriptsize \color{gray} 0.0}& 3.4{\scriptsize \color{gray} 0.2}\\ 
Lab\_2 & 2 & 3 & 23.0{\scriptsize \color{gray} 0.0}& 2.1{\scriptsize \color{gray} 0.1}& 31.0{\scriptsize \color{gray} 0.0}& 2.2{\scriptsize \color{gray} 0.1}& 27.0{\scriptsize \color{gray} 0.0}& 2.1{\scriptsize \color{gray} 0.1}& 25.0{\scriptsize \color{gray} 0.0}& 2.1{\scriptsize \color{gray} 0.1}\\ 
Lab\_3 & 4 & 5 & 25.0{\scriptsize \color{gray} 0.0}& 3.4{\scriptsize \color{gray} 0.3}& 50.6{\scriptsize \color{gray} 1.1}& 4.2{\scriptsize \color{gray} 0.3}& 42.0{\scriptsize \color{gray} 0.0}& 3.4{\scriptsize \color{gray} 0.2}& 30.0{\scriptsize \color{gray} 0.0}& 3.4{\scriptsize \color{gray} 0.2}\\ 
Lab\_4 & 4 & 9 & 71.8{\scriptsize \color{gray} 4.9}& 6.9{\scriptsize \color{gray} 0.6}& 135{\scriptsize \color{gray} 0.0}& 7.9{\scriptsize \color{gray} 0.5}& 164{\scriptsize \color{gray} 3.8}& 7.6{\scriptsize \color{gray} 0.5}& 110{\scriptsize \color{gray} 6.8}& 6.5{\scriptsize \color{gray} 0.3}\\ 
Lab\_5 & 12 & 17 & 93.0{\scriptsize \color{gray} 0.0}& 20.4{\scriptsize \color{gray} 0.9}& 285{\scriptsize \color{gray} 0.0}& 27.0{\scriptsize \color{gray} 1.2}& 278{\scriptsize \color{gray} 0.0}& 20.5{\scriptsize \color{gray} 1.4}& 207{\scriptsize \color{gray} 0.0}& 14.3{\scriptsize \color{gray} 0.7}\\ 
Field\_1 & 2 & 4 & 16.0{\scriptsize \color{gray} 0.0}& 3.2{\scriptsize \color{gray} 1.1}& 27.0{\scriptsize \color{gray} 5.6}& 3.6{\scriptsize \color{gray} 0.4}& 23.0{\scriptsize \color{gray} 0.0}& 3.1{\scriptsize \color{gray} 0.9}& 16.0{\scriptsize \color{gray} 0.0}& 2.5{\scriptsize \color{gray} 0.2}\\ 
Field\_2 & 2 & 6 & 53.0{\scriptsize \color{gray} 0.0}& 6.2{\scriptsize \color{gray} 0.3}& 99.0{\scriptsize \color{gray} 0.0}& 12.9{\scriptsize \color{gray} 0.5}& 75.0{\scriptsize \color{gray} 0.0}& 6.5{\scriptsize \color{gray} 0.3}& 56.0{\scriptsize \color{gray} 0.0}& 6.1{\scriptsize \color{gray} 0.3}\\ 
Field\_3 & 4 & 8 & 84.0{\scriptsize \color{gray} 0.0}& 12.6{\scriptsize \color{gray} 0.8}& 226{\scriptsize \color{gray} 26.9}& 27.6{\scriptsize \color{gray} 2.1}& 141{\scriptsize \color{gray} 4.9}& 12.8{\scriptsize \color{gray} 1.1}& 138{\scriptsize \color{gray} 0.0}& 13.0{\scriptsize \color{gray} 0.8}\\ 
Field\_4 & 6 & 10 & 77.0{\scriptsize \color{gray} 0.0}& 13.4{\scriptsize \color{gray} 0.7}& 247{\scriptsize \color{gray} 1.9}& 30.5{\scriptsize \color{gray} 1.2}& 195{\scriptsize \color{gray} 0.0}& 14.5{\scriptsize \color{gray} 1.0}& 188{\scriptsize \color{gray} 0.0}& 14.5{\scriptsize \color{gray} 0.6}\\ 
Field\_5 & 6 & 11 & 289{\scriptsize \color{gray} 0.0}& 50.3{\scriptsize \color{gray} 1.1}& 842 & 103 & 739{\scriptsize \color{gray} 56.6}& 57.5{\scriptsize \color{gray} 8.5}& 573{\scriptsize \color{gray} 1.1}& 46.4{\scriptsize \color{gray} 0.9}\\ 
\midrule 
total & & & 1672 & 283 & 3526 & 404 & 3766 & 305 & 2743 & 235 \\ 
\bottomrule  \\
\end{tabular} 
\caption{Analysis of different relaxations in TAMP. Number of NLP evaluations and CPU time, averaged over 10 random seeded runs, with standard deviations in gray. ``--'' indicates consistent failure to solve a problem within 100 seconds. 
}
\label{table:relaxations} 
\end{table*}

We analyze in detail the 4 relaxations 
to detect minimal conflicts in TAMP problems (see Sec. \ref{sec:relax}):

\begin{itemize}
  \item Removal of trajectories (\emph{No Traj})
  \item Removal of collision constraints (\emph{No Col})
  \item Removal of time consistency (\emph{No Time})
  \item Removal of robots variables (\emph{No Robot})
\end{itemize}

In the experimental evaluation, our algorithms \textit{GNPP\_tr}, \textit{GNPP\_trn}, and \textit{ GNPP\_trng} check the following relaxation rules before solving the complete graph-NLP.

\begin{enumerate}
  \item \emph{No Robot} + \emph{No Col} + \emph{No Traj}
  \item \emph{No Col} + \emph{No Traj}
  \item \emph{No Robot} + \emph{No Time} + \emph{No Traj} 
  \item \emph{No Time} + \emph{No Traj}
  \item \emph{No Traj} 
\end{enumerate}


Namely, we first apply (1) to the original graph-NLP and check if it is feasible (breaking the full graph 
into disconnected components, and doing binary search on the time index, see Alg. \ref{alg:infeas}). If feasible, apply (2) to the original graph and check if it feasible. If feasible, apply (3) and so forth.
The algorithm stops the first time it finds an infeasible subgraph. 
If all tested relaxations are feasible, we solve for the full graph-NLP.

To analyze the influence of each individual relaxation we evaluate the following relaxation rules:

\begin{itemize}
  \item
    Without \emph{No Col}  (\textit{GNPP\_tnr1})

\begin{enumerate}
  \item \emph{No Robot} + \emph{No Time} + \emph{No Traj}
  \item \emph{No Robot} + \emph{No Traj}
  \item \emph{No Time} + \emph{No Traj}
  \item \emph{No Traj}
\end{enumerate}


\item Without \emph{No Time} (\textit{GNPP\_tnr2})


\begin{enumerate}
  \item \emph{No Robot} + \emph{No Col} + \emph{No Traj}
  \item \emph{No Robot} + \emph{No Traj}
  \item \emph{No Col}  + \emph{No Traj}
  \item \emph{No Traj} 
\end{enumerate}


\item Without \emph{No Robot} (\textit{GNPP\_tnr3})

\begin{enumerate}
  \item \emph{No Col} + \emph{No Time} + \emph{No Traj}
  \item \emph{No Col} + \emph{No Traj}
  \item \emph{No Time} + \emph{No Traj}
  \item \emph{No traj}
\end{enumerate}


\end{itemize}

Results are shown in Table \ref{table:relaxations}.
We use the version of our algorithm called \textit{GNPP\_trn} (with relaxation rules \textit{``r"}, search on the time index \textit{``t"}, and convergence heuristic \textit{``n"}). \textit{GNPP\_tnr0} is the default implementation using all the relaxations. \textit{GNPP\_tnr1}, \textit{GNPP\_tnr2}, and \textit{GNPP\_tnr3} apply three different relaxations rules (see above).

The worst performing variation is \textit{GNPP\_trn1} (without \textit{No Col}). This highlights that \emph{No col} relaxation is fundamental to detect small conflicts in TAMP, as it makes the graph-NLP sparse and, when combined with the search on the time index, highly disconnected. 
The best performing variation is \textit{GNPP\_trn3}, suggesting that \emph{No Robot} relaxation is in fact not useful to improve the running time.
Finally, we remark that the overall performance of each variation is influenced by the \textit{convergence heuristic}, that potentially reduces the size of the infeasible subgraph and outputs small conflicts even when no informative relaxation is used.


Note that the relaxation \textit{No Traj} is applied in all the relaxation rules, and we only 
compute trajectories if all the previous relaxations are found to be feasible (that is, when solving for the full graph-NLP). The justification is twofold: first, the nonlinear optimization problem with trajectories contains 20 times more scalar variables than the optimization without trajectories,
and is an order of magnitude slower to solve (each trajectory is represented with 20 waypoints, while a mode-switch is represented with a single waypoint). Second, a good strategy to solve the optimization problem including the trajectories is to 1) compute the mode-switches, 2) warmstart the trajectories with a linear interpolation between the mode-switches, and 3) reoptimize trajectories and mode-switches jointly.


\subsection{Extended Study of Scalability: Objects and Obstacles}
\label{ref:extended-scale}


In this section, we study the scalability of our solver when increasing the number of objects and obstacles. These results extend the benchmark in the Experimental Results (Sec. \ref{sec:experimental_results} and Table \ref{table:thetable}). 

\begin{figure}
  \centering

  \includegraphics[trim=0 0 0 2cm,clip,width=.45\linewidth ]{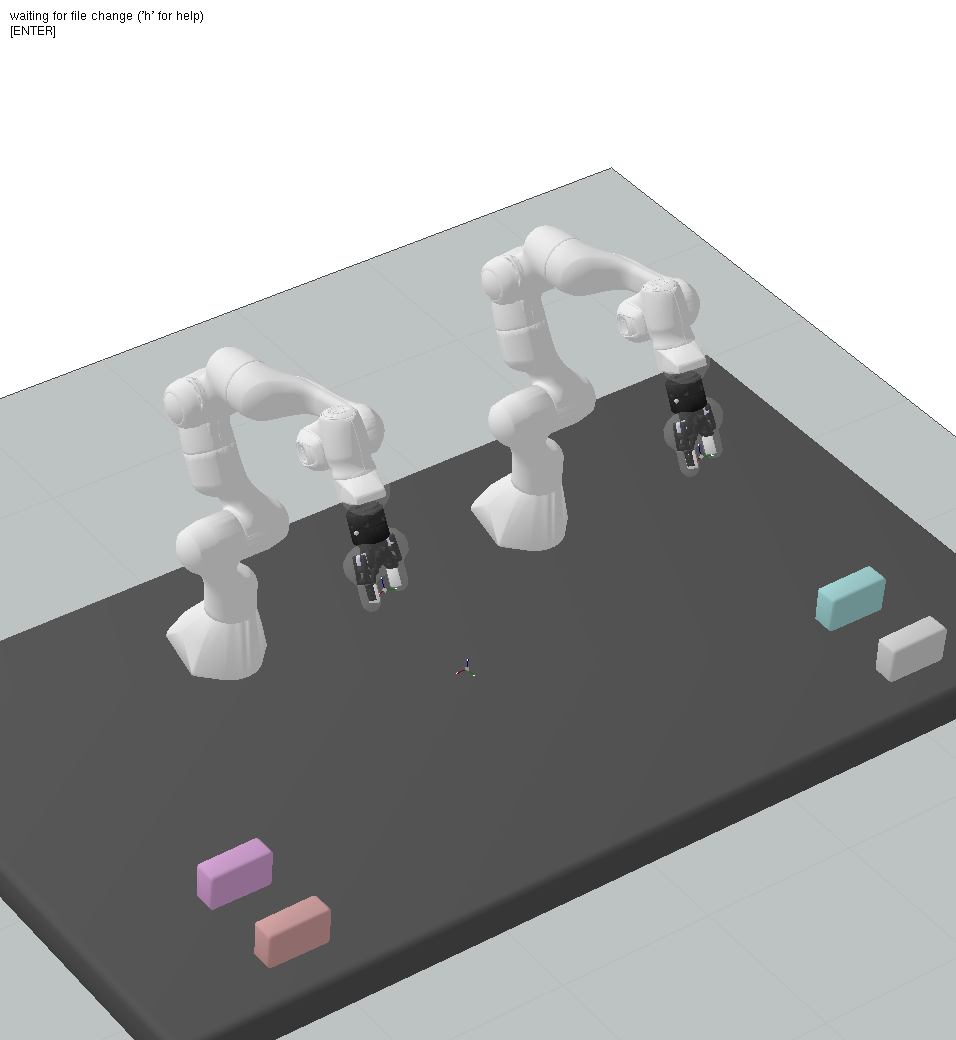}
  \includegraphics[trim=0 0 0 2cm,clip,width=.45\linewidth]{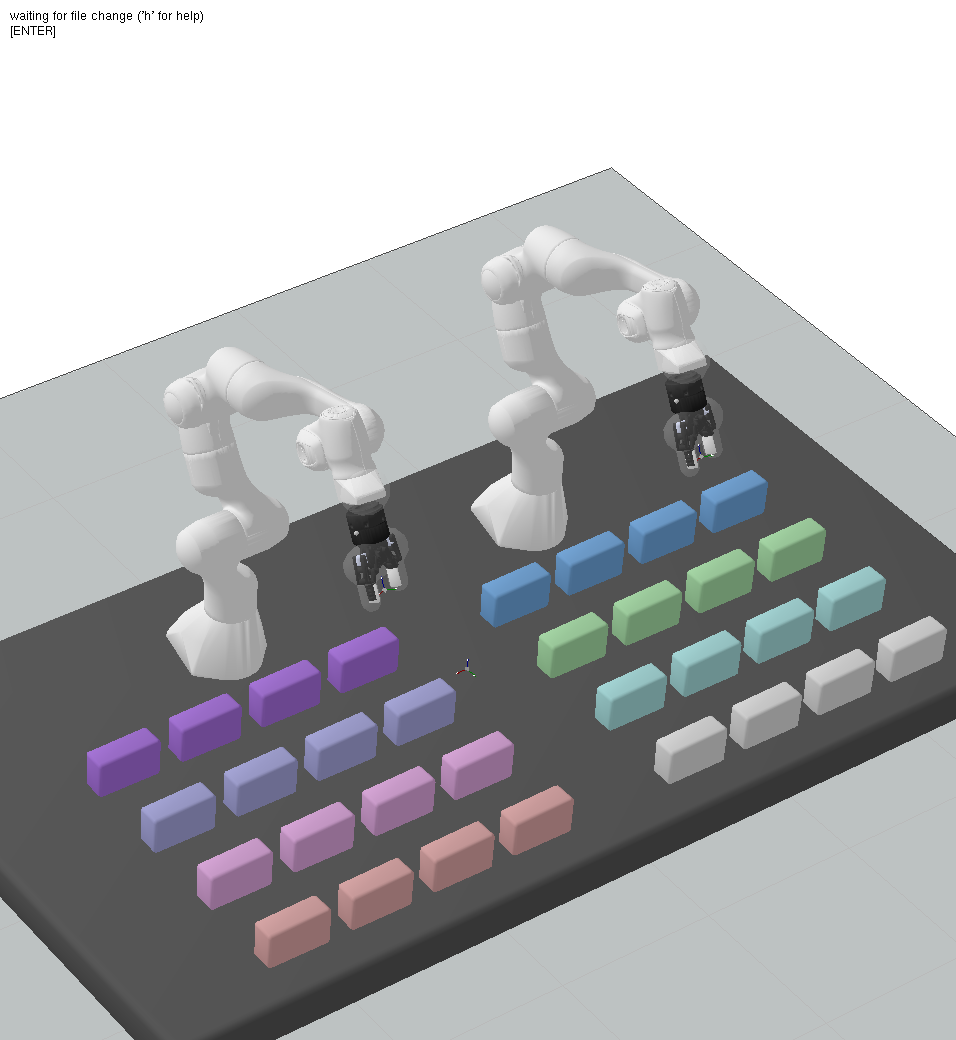}

  \caption{\emph{Stacking boxes}. The goal is to stack blocks in pairs, e.g. put the orange block on top of the pink and the white on top of the blue block. \textit{Left}: 4 objects. \textit{Right}: 32 objects.} \label{fig:scale_objects} \end{figure}

\begin{figure}
  \centering
  \includegraphics[trim=0 0 0 2cm,clip,width=.45\linewidth]{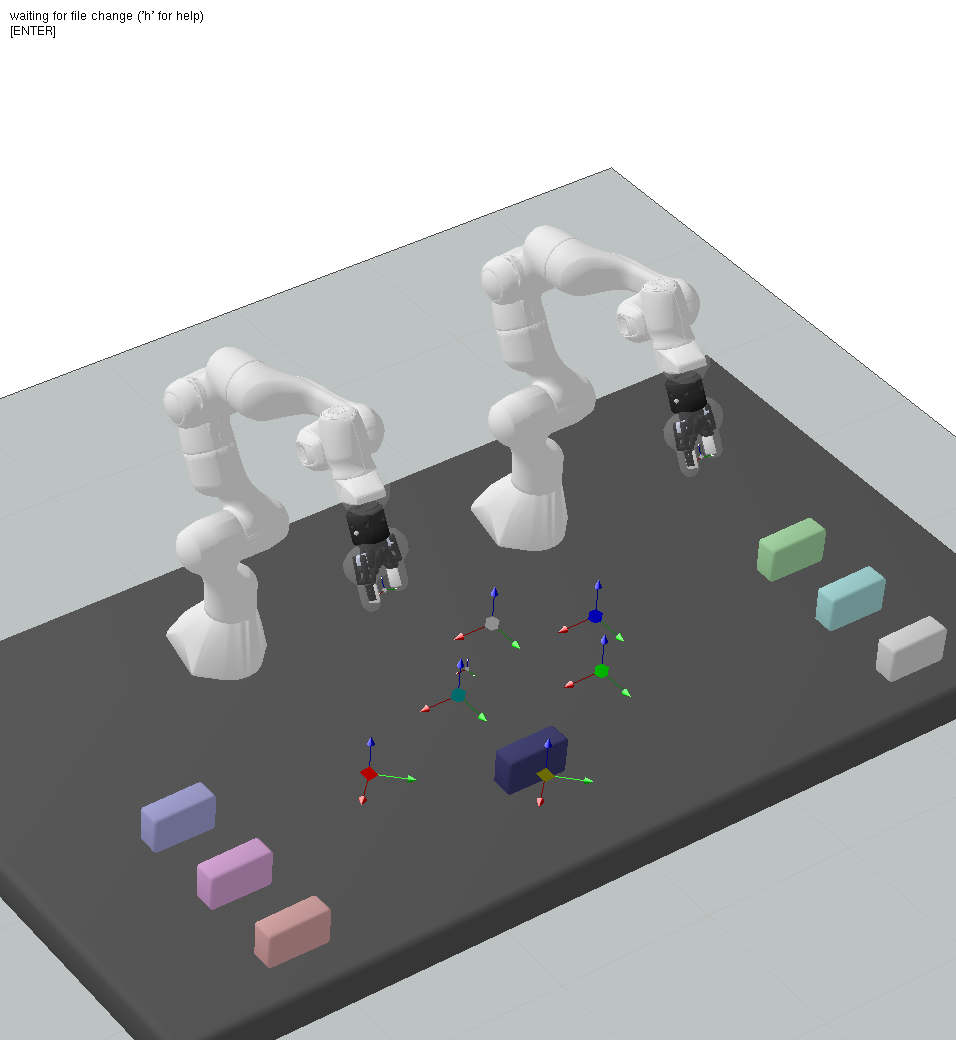}
  \includegraphics[trim=0 0 0 2cm,clip,width=.45\linewidth]{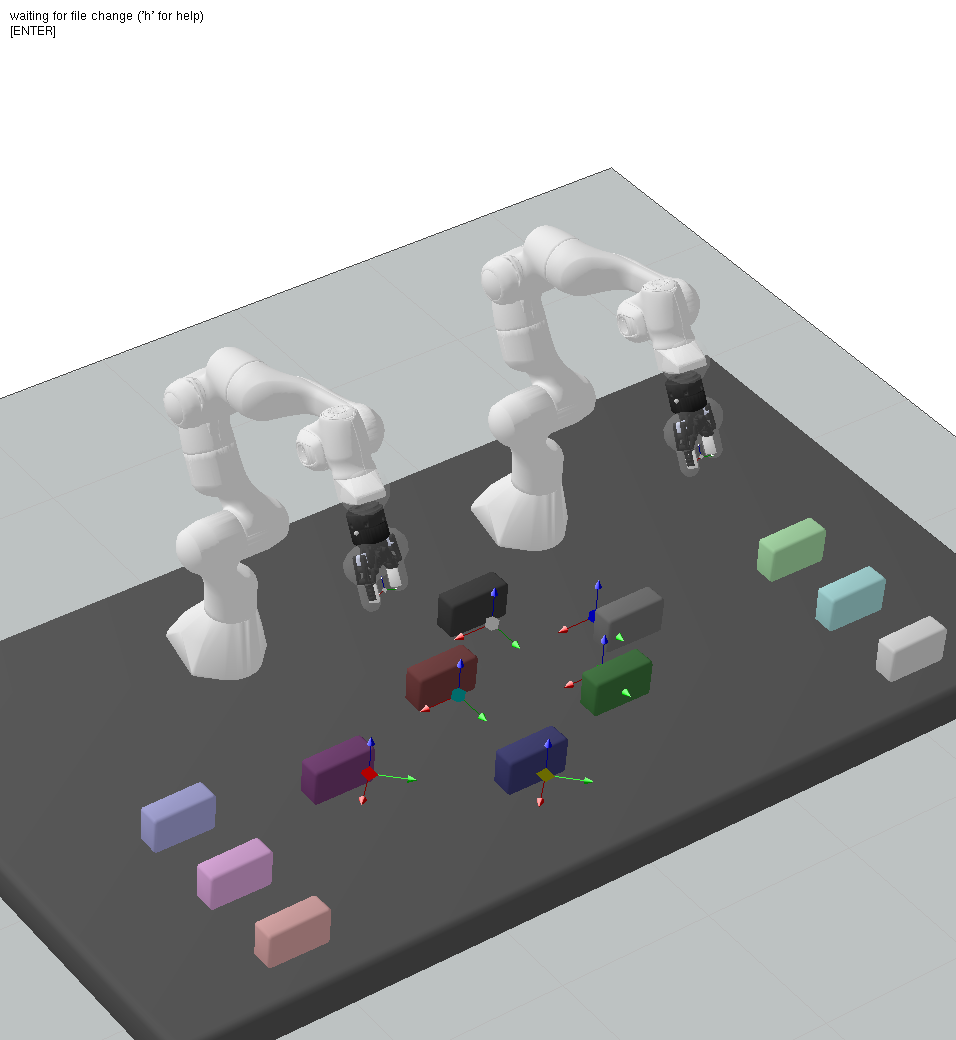}
  \caption{\emph{Placement in a cluttered table}. Dark colors denote movable obstacles. 
    The frame markers show the goal positions and orientations for the objects.
  \textit{Left}: 1 obstacle. \textit{Right}: 6 obstacles.}
    \label{fig:scale_obs}
\end{figure}

The new problems are based on the \textit{Laboratory} scenario.
\begin{itemize}
  \item \textit{Stacking Boxes}. 
    Two 7-DOF manipulator arms execute pick and place actions to stack blocks
    in small towers of two blocks. We increase the number of blocks in each instance, from 4 to 32. See Fig. \ref{fig:scale_objects}.
  \item \textit{Placement in a cluttered table}.
    Two 7-DOF manipulator arms execute pick and place actions to place 6 blocks into a goal position and orientation, which requires to detect possible collisions and move obstacles around the table. We increase the number of movable obstacles from 1 to 6. See Fig. \ref{fig:scale_obs}.
\end{itemize}

\begin{figure}[t]
  \centering
  \includegraphics[width=.6\linewidth]{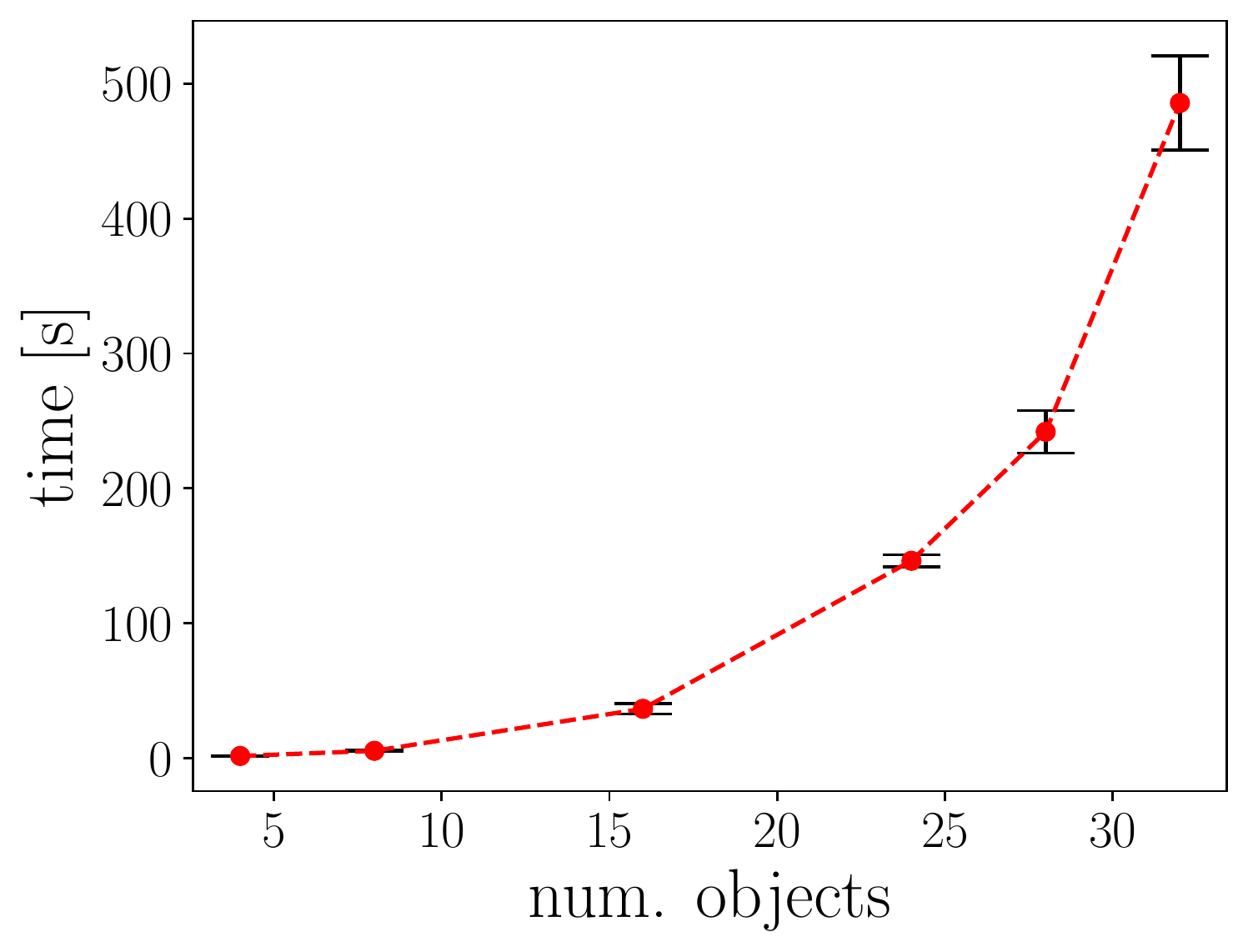}
  \caption{Computational time in \textit{Stacking boxes}. See Fig. \ref{fig:scale_objects}. We report mean and standard deviation over 10 runs.}
  \label{fig:plot_obj}
\end{figure}

\begin{figure}[t]
  \centering
  \includegraphics[width=.6\linewidth]{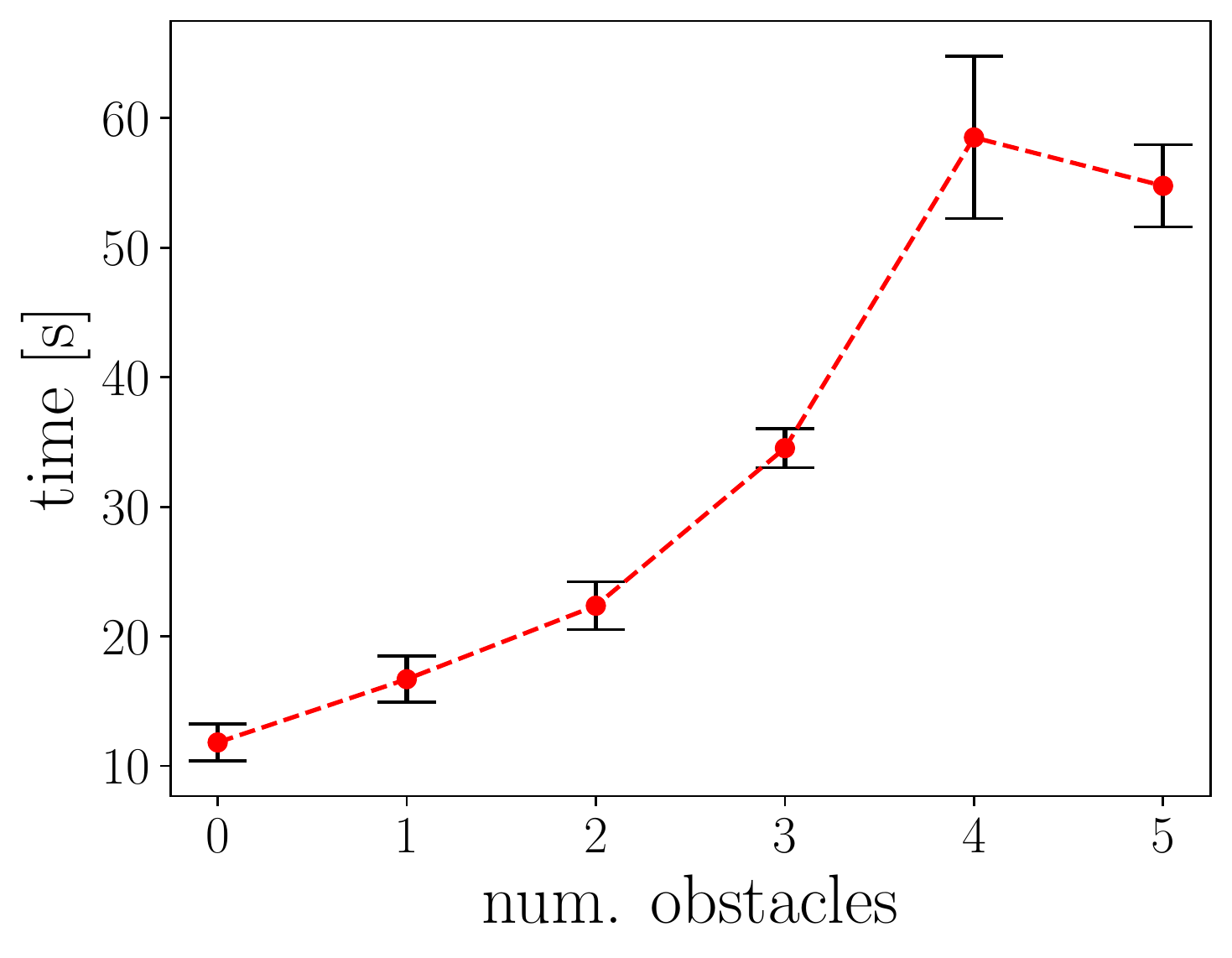}
  \caption{Computational time in \textit{Placement in a cluttered table}. See Fig. \ref{fig:scale_obs}. We report mean and standard deviation over 10 runs.}
  \label{fig:plot_obs}
\end{figure}

Results are shown in Fig. \ref{fig:plot_obs} and Fig. \ref{fig:plot_obj}.

\textit{Stacking Boxes} - 
Our solver scales polynomially to the number of objects in the scene, while relying on joint optimization and not using hand-crafted problem decompositions or sampling of partial solutions.
The number of evaluated action plans scales linearly with the number of objects, but the computational time spent on solving the (sub)graph-NLPs increases polynomially with the size of the nonlinear optimization problem.
In fact, the largest problem requires a motion plan of 32 actions. In this setting, generating a full motion using joint optimization is not the most efficient approach, and could be improved with an heuristic strategy that fixes mode-switches (after joint optimization) and solves the trajectory individually for each step.

\textit{Placement in a cluttered table} - In this setting, our solver scales linearly with the number of obstacles. This is achieved by the efficient detection and encoding of minimal infeasible subgraphs (in this case, which obstacles are blocking the placement of a block). 

Based on the results of the main benchmark (Tab. \ref{table:thetable}) and the scalability study (Fig. \ref{fig:plot_obj} and \ref{fig:plot_obs}), we can conclude that the running time of our solver depends on:

\begin{itemize}
  \item \textit{Solving the Logic Problem}: fast in practice using the logic encoding of geometric conflicts and a state-of-the-art PDDL planner (the worst case time complexity is exponential in the action branching factor).
  \item \textit{Number of iterations of GNPP}: the efficient detection and encoding of geometric information achieves practical linear  complexity (the worst case is exponential in the action branching factor).
  \item \textit{The number of evaluated subgraphs}: for each new symbolic sequence, the algorithm to detect infeasible subgraphs evaluates a number of subgraphs that is linear in the number of objects (in practice) and logarithmic in the length of the action sequence. Worst case is exponential in the number of objects.
  \item \textit{Solving nonlinear programs}: the theoretical time complexity (assuming a bounded number of nonlinear iterations) is cubic in the number of objects and robots and linear in the length of the action sequence. The practical complexity is worse, because the nonlinear optimizer usually requires more iterations to solve larger NLPs.
\end{itemize}


\end{document}